%% file: main.tex
\documentclass[conference]{IEEEtran}
\IEEEoverridecommandlockouts
\usepackage{cite}
\usepackage{amsmath,amssymb,amsfonts}
\usepackage{algorithmic}
\usepackage{graphicx}
\usepackage{textcomp}
\usepackage{xcolor}
\usepackage{subcaption}
\usepackage{url}
\usepackage{multirow}
 \usepackage{booktabs}
 \usepackage[symbol]{footmisc}

\def\BibTeX{{\rm B\kern-.05em{\sc i\kern-.025em b}\kern-.08em
    T\kern-.1667em\lower.7ex\hbox{E}\kern-.125emX}}
\begin{document}

\title{Contrastive Graph Convolutional Networks for \\Hardware Trojan Detection in Third Party IP Cores}

\author{\IEEEauthorblockN{Nikhil Muralidhar\IEEEauthorrefmark{1}\IEEEauthorrefmark{4}\thanks{\IEEEauthorrefmark{4}Corresponding Author Email: nik90@vt.edu},
Abdullah Zubair\IEEEauthorrefmark{2}, Nathanael Weidler\IEEEauthorrefmark{3}, Ryan Gerdes\IEEEauthorrefmark{2} and
Naren Ramakrishnan\IEEEauthorrefmark{1}}\vspace{0.3cm}
\IEEEauthorblockA{\IEEEauthorrefmark{1}Department of Computer Science,
Virginia Tech,
Arlington, VA\\
\IEEEauthorrefmark{2}Department of Electrical and Computer Engineering, Virginia Tech, Arlington, VA\\
\IEEEauthorrefmark{3}Space Dynamics Laboratory, Logan, UT\\
}}
 
\iffalse 
\author{\IEEEauthorblockN{Nikhil Muralidhar}
\IEEEauthorblockA{\textit{Dept. of Computer Science} \\
\textit{Virginia Tech}\\
Arlington, Virginia \\
nik90@vt.edu}
\and
\IEEEauthorblockN{Abdullah Zubair}
\IEEEauthorblockA{\textit{Dept. of Electrical \& Computer Engineering} \\
\textit{Virginia Tech}\\
Arlington, Virginia \\
abdullahzubair@vt.edu}
\and
\IEEEauthorblockN{Nathanael Weidler}
\IEEEauthorblockA{
\textit{Space Dynamics Laboratory}\\
Logan, Utah \\
nweidler@gmail.com}
\and
{\IEEEauthorblockN{\hspace{2.5cm}Ryan Gerdes}
\IEEEauthorblockA{\hspace{2.5cm}\textit{Dept. of Electrical \& Computer Engineering} \\
\textit{\hspace{2.5cm}Virginia Tech}\\
\hspace{2.5cm}Arlington, Virginia \\
\hspace{2.5cm}rgerdes@vt.edu}}
\and
\IEEEauthorblockN{Naren Ramakrishnan}
\IEEEauthorblockA{\textit{Dept. of Computer Science} \\
\textit{Virginia Tech}\\
Arlington, Virginia \\
naren@cs.vt.edu}
}
\fi 
\maketitle

\newcommand{\nikhilc}[1]{\textcolor{red}{Nikhil says: #1}}
\newcommand{\abdullahc}[1]{\textcolor{green}{#1}}
\newcommand{\narenc}[1]{\textcolor{orange}{Naren says: #1}}
\newcommand{\ryanc}[1]{\textcolor{blue}{Ryan says: #1}}
\newcommand{\ourmethodfull}{Graph-Aware Trigger Detection Network\,}
\newcommand{\ourmethod}{GATE-Net\,}

\begin{abstract}
The availability of wide-ranging third-party intellectual property (3PIP) cores enables integrated circuit (IC) designers to focus on designing high-level features in ASICs/SoCs. The massive proliferation of ICs brings with it an increased number of bad actors seeking to exploit those circuits for various nefarious reasons. This is not surprising as integrated circuits affect every aspect of society.
Thus, malicious logic (Hardware Trojans, HT) being surreptitiously injected by untrusted vendors into 3PIP cores used in IC design is an ever present threat.
In this paper, we explore methods for identification of trigger-based HT in designs containing synthesizable IP cores without a golden model. Specifically, we develop methods to detect hardware trojans by detecting triggers embedded in ICs purely based on netlists acquired from the vendor.
We propose~\ourmethod, a deep learning model based on graph-convolutional networks (GCN) trained using supervised contrastive learning, for flagging designs containing randomly-inserted triggers using only the corresponding netlist. Our proposed architecture achieves significant improvements over state-of-the-art learning models yielding an average 46.99\% improvement in detection performance for combinatorial triggers and 21.91\% improvement for sequential triggers across a variety of circuit types. Through rigorous experimentation, qualitative and quantitative performance evaluations, we demonstrate effectiveness of \ourmethod and the supervised contrastive training of \ourmethod for HT detection.

\iffalse 
  Proliferation of untrusted third-party IP vendors carries a greater risk of hardware trojans being surreptitiously injected into the circuit\fi  
\end{abstract}
\begin{IEEEkeywords}
hardware trojan, machine learning, graph convolutional network, deep learning, contrastive learning
\end{IEEEkeywords}

\section{Introduction}\label{sec:introduction}
\input{introduction}

% \section{Related Work}\label{sec:related_work}
% \input{related_work}
\section{Generation and Representation of Circuits }\label{sec:background}
\input{background}
\section{HT Trigger Detection with~\ourmethod}\label{sec:problem_formulation}
\input{problem_formulation}
\section{Experimental Setup and Evaluation}\label{sec:expt_setup}
\input{experimental_setup}
\section{Results and Discussion}\label{sec:results}
\input{results_and_discussion}
\section{Conclusion}\label{sec:conclusion}
\input{conclusion}

\bibliographystyle{IEEEtran}
\bibliography{bibfile}
\end{document}

%% file: introduction.tex
%\ryanc{Side-channel approaches, e.g., are not applicable}

 Hardware Trojans (HTs) are malicious modifications to an integrated circuit (IC), which can change their intended functionality or cause them to malfunction. Parametric Trojans modify the existing logic of the circuit while functional Trojans are realized by adding transistors and gates to the circuit. HTs are also classified as always-on or triggered based on the mechanism of activation. Once activated, the Trojans alter the reliability, leak confidential information such as secret keys, or cause an IC to malfunction. They can cause an IC chip to become disabled or compromised, allowing an adversary to gain access to highly protected data~\cite{tehranipoor2010survey}. 

The demand for high performance, low cost and multi-functional ICs continues to rise worldwide. To keep up with these demands, and to stand-out in the highly competitive market, chip designers, big corporations, and small start-ups alike either seek third-party intellectual property (3PIP) cores or employ third-party design houses to outsource standard and commonly used logic designs. Although this helps them focus on the novel features of their design, it significantly increases risk of a HT being implanted in the chip, as untrusted 3PIP vendors, or third party contractors working on the design, could surreptitiously insert a Trojan into the design. Even if the vendors and contractors are trustworthy, compromised design tools could implant HTs into  a circuit without their knowledge \cite{huang2020survey}. A golden IC, that is, a verified Trojan-free version of an IC, is often used as a reference to detect Trojans using comparative analysis \cite{li2015survey, agrawal2007trojan, chakraborty2009mero, chakraborty2009hardware,courbon2015high}, e.g., ensuring that side-channels are consistent between the golden model and a chip for which we wish to establish the presence or absence of a HT. By their very nature 3PIP cores preclude the existence of golden models (as 3PIP is opaque to the user and its integrity can only be attested to by the potentially untrustworthy/compromised vendor), thus preventing the use of HT detection approaches based on golden models.  

We present a methodology for applying machine learning (ML) models (specifically graph-convolutional networks, GCN) trained with supervised contrastive learning~\cite{khosla2020supervised} for improved representation learning on the task of HT detection, without relying on a golden model, with the goal of detecting Trojans in ICs inserted prior to chip fabrication. We are interested in HT that are stealthy (i.e., do not affect IC operation until called upon by the attacker to do so) and may be easily activated by the attacker.  To this end we focus on detecting HT that are triggered based upon input to the IC. \iffalse A deep learning application requires a large amount of data in a form it can digest.  We solve this problem through representing the gate-level circuit in a compact matrix form, known as, circuit adjacency matrix and embedding synthetic HT triggers.\fi We show promising results applying learning strategies to detect combinatorial and sequential HT triggers comprised of standard cells such as logic gates and flip-flops. We find that GCN models trained on circuit netlists with cell types as features are surprisingly effective at identifying circuits with embedded Trojan triggers. We find that GCN models perform significantly better than state-of-the-art baseline classifiers.

\subsection{Contributions}
We propose \ourmethodfull{} (\ourmethod{}), a novel GCN-based supervised contrastive representation learning model for learning salient features of HT-embedded vs.\ HT-free circuits using only the gate-level netlist of the design.
\iffalse \begin{enumerate}
    \item ~\ourmethod does not rely on a golden model for trigger-based HT detection.
    \item We present a rigorous analysis of the performance of~\ourmethod and compare it with several state-of-the-art classification models for trigger-based HT detection.
    \item We present a novel methodology to create datasets for HT detection research. Our dataset, with combinatorial and sequential trigger inserted circuits, in adjacency matrix form will be made publicly available.
\end{enumerate}\fi 
\begin{enumerate}
\item  ~\ourmethod does not rely on a golden model for trigger-based HT detection.
\item  We are the first to employ Graph Convolutional Networks as a methodology to model hardware trigger detection using unstructured circuit data.
\item  We present a rigorous analysis of the performance of~\ourmethod and compare it with several state-of-the-art classification models for trigger-based HT detection.
\item  We present a novel methodology to create datasets for HT detection research. Our dataset, with combinatorial and sequential trigger inserted circuits, in adjacency matrix form will be made publicly available along with our source code for the proposed \ourmethod model, trigger generation and embedding.
\end{enumerate}
\iffalse 
\begin{enumerate}
    \item We propose~\ourmethodfull (\ourmethod) a novel GCN based representation learning model for learning salient features of HT-embedded vs. HT-free circuits using only gate-level netlists.~\ourmethod does not rely on a golden model for trigger-based HT detection.
    \item We present a rigorous analysis of the performance of~\ourmethod and compare it with several state-of-the-art classification models for trigger-based HT detection.
    \iffalse 
    \item We propose a novel problem of identifying HT triggers instead of net classification as opposed to previous approaches.\nikhilc{Abdullah: why is predicting the trigger better?} \abdullahc{need to refer cited papers in more detail to answer}
    \fi 
    \item We present a novel methodology to create datasets for HT detection research.
    
    \item Our dataset with combinatorial and sequential trigger inserted circuits in adjacency matrix form is publicly available.

    %\item We also propose a dataset generation routine for circuits embedded with trigger-based HT in circuit adjacency matrix form, ideal for deep learning consumption. The dataset comprising of more than 70 IP cores embedded with triggers is made publicly available.
\end{enumerate}
\fi 
\subsection{Related Work}
%\nikhilc{Abdullah: Please expand the related work to include approaches we discussed recently.}
\par\noindent
Side-channel analysis has been used to detect HTs at the circuit level by analyzing the power characteristics or timing delays in the infected circuit and comparing with that of a golden model\cite{tehranipoor2010survey}. The HT detection methods based on boolean function analysis on the circuit has also been proposed \cite{waksman2013fanci} \cite{zhang2015veritrust}. These methods rely on the combinatorial logic of the circuits and therefore have limitations against the sequential HTs and become computationally expensive with increase in circuit size.

Recently, ML and deep learning models have shown promise in various  fields  like  computer  vision,  natural  language  processing (NLP) and  time  series  analysis and are increasingly being applied to HT detection. The structural and functional features derived from the gate-level representation of the circuits provide crucial data for learning models in detecting HTs. Yao et al. \cite{yao2015fastrust}, in their proposed method FASTrust, performed structural feature analysis on the flow graphs created from the flipflops in the gate-level representation and Chen et al. \cite{chen2017hardware} proposed ML-FASTrust by including the functional features of the combinatorial logic in addition to the structural features.
 In~\cite{hasegawa2016hardware,hasegawa2017Trojan,kurihara}, the authors develop specific \textit{structural} features, based on proximity to specific circuit components like flip-flops and also the count of nets (connection between two logic components) k-hops away from a particular net, in order to classify each net in a netlist as either a \textit{Trojan-net} or a \textit{normal-net} using support-vector machines (SVM) and Random Forest (RF) for classification. A gradient boosting classifier for HT detection is proposed in~\cite{han2019hardware}. The classifier operates on features extracted from an abstract syntax tree (AST) derived form the netlist. \iffalse The AST is then processed using specific feature extractors which look for features corresponding to proportion of specific operation types (e.g., \textit{if}, \textit{always} statements).\fi In~\cite{oya2015score} a score-based mechanism is proposed for identifying HTs using 10 circuit component specific features used to provide weak supervision for the classification task. Shen et al. \cite{shen2017lmdet} propose LMDet, an NLP based approach for detecting HTs using n-gram sequencing. Hoque et al. \cite{hoque18} uses ensemble of three different ML models, namely RF, Naive Bayes and Adaptive Boosting over a set of structural and functional features.

  The aforementioned methods do not explicitly operate on the entirety of the original graph domain obtained from the netlist. Functional features (and a limited set of structural features) in general may fail to capture the rich knowledge present in the circuit topology. GNN4TJ \cite{yasaei2021gnn4tj} is a recent model that uses data-flow-graph derived from RTL code for HT detection. In the semiconductor supply-chain, GATE-Net (which works with gate-level netlists) will capture HTs injected at all stages of design while GNN4TJ requires additional expensive reverse-engineering for HTs injected after Logic Synthesis. 
  From the above-mentioned approaches, we implement the methods proposed by Kurihara et al.\cite{kurihara} and Hoque et al.\cite{hoque18} and evaluate \ourmethod against them in Section \ref{sec:results}.
 
\iffalse  Finally, a hardware circuit representation learning method is proposed in~\cite{daiconvcirc2017} based on simple logical convolution and pooling of node representations based on the logical-\texttt{OR} operation between the representation of a node and the representations of its immediate neighbors. We employ this method as a baseline to compare with~\ourmethod for the HT trigger detection task.\fi 

\textbf{Hardware Trojan Dataset}: One of the challenges associated with conducting research in the field of HTs is the relative lack of publicly available Trojan-inserted circuits. This is particularly true when attempting to apply ML strategies which require large amounts of data for effective model training. The Trust-Hub benchmark \cite{trustHub} contains 96 Trojan-inserted circuits, sub-categorized by multiple taxonomies,  such as trojan location and activation mechanism. Our criteria for HT selection is based on stealthiness and ease of activation; i.e., the HT must use a trigger that is activated via common, user-facing IC inputs. As only five of the circuits in the Trust-Hub benchmark contain trigger-based HT that are activated based on legitimate user input, we introduce (Section~\ref{sec:background}) a method to generate trigger-based HT using the gate-level netlist of publicly available IP cores (Section~\ref{subsec:dataset}).
\iffalse In this paper, we introduce a dataset generation methodology that takes in a synthesized, gate-level netlist of a circuit and generates a circuit adjacency matrix. The circuit adjacency matrix, the formal definition of which is addressed in the next section, has the complete information of the circuit in a compact form and is effective at training ML models. The triggers, also in the adjacency matrix form, are then inserted in the benign circuit's matrix to be used for HT detection techniques. The data generation is detailed in Section~\ref{subsec:dataset}.\fi
%\par\noindent

%\textbf{ML for Hardware Trojan Detection}:

\subsection{Learning Approach Introduction}
The aforementioned methods are primarily based on specific feature engineering, hence a more general representation learning method which automatically extracts IC features is necessary to enable a generalizable HT detection framework. Neural networks have been effective at representation learning without explicit feature engineering. Recently, Graph Convolutional Networks (GCN)~\cite{kipf2016semi}, a variant of deep neural networks capable of directly operating on irregular grids like graphs, have been successfully used across various disciplines like studying chemical compounds and molecular structure, \iffalse to learn molecular fingerprints, predict molecular properties and to synthesize chemical compounds. GCNs have also been used \fi in computer vision for scene generation, point cloud classification, for text classification in natural language processing. \iffalse and for computer program verification.\fi The various successes of GCNs are detailed in a recent comprehensive survey~\cite{wu2020comprehensive}. We employ a GCN based learning architecture~\ourmethod for effective representation learning of hardware circuit features \emph{automatically}, sans any manual feature engineering for HT trigger detection in ICs. In order to improve the representations learnt by GCNs we also employ a supervised variant of a recently popular representation learning approach called \emph{contrastive learning}~\cite{chen2020simple,you2020graph,jaiswal2021survey} and through extensive experiments demonstrate the effectiveness of GCNs combined with supervised contrastive learning for improved performance on HT detection.

\subsection{Threat Model}
%\ryanc{Access to netlist and where attacker can corrupt.}
We assume that malicious actors have implanted a trigger-based HT into an IC during the design phase. The malicious actors can be either one (or a combination) of the following: a) an untrusted 3PIP core vendor who implants the Trojan in the IP core purchased by the designer to be used in the IC, b) a compromised design tool, used by 3PIP vendor or designer, c) an untrusted third party designer to whom the design is outsourced, or d) a malicious designer on the team from inside the organization. The 3PIP cores considered in the threat model are the soft (synthesizable) IP cores and it is assumed that the gate-level netlist of the final circuit design is available to the chip designer (defender).

\begin{figure}[tb]
\centering
\begin{subfigure}[b]{\columnwidth}
   \centering
   \includegraphics[scale=0.7]{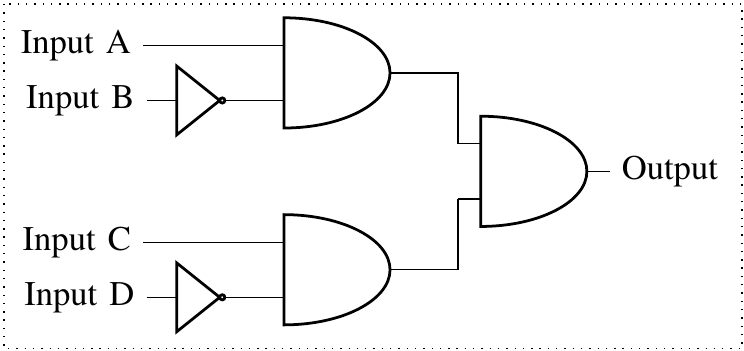}
   \caption{}
   \label{fig:Ng1} 
\end{subfigure}
\vfill
\begin{subfigure}[b]{\columnwidth}
   \centering
   \includegraphics[scale=0.55]{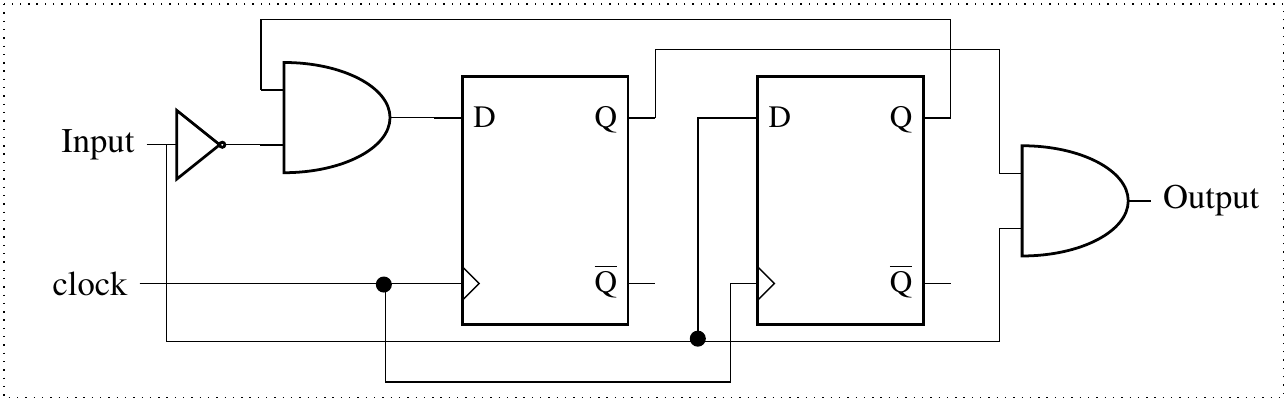}
   \caption{}
   \label{fig:Ng2}
\end{subfigure}
\caption[Triggers]{(a) A 4-input combinatorial trigger example (b) A sequential trigger example ('101' sequence detector) }
\label{fig:trigger}
\end{figure}

The proposed approach is independent of the functionality of the HT once it is triggered, only that it is latent until triggered. The trigger can be either combinatorial or sequential. A combinatorial trigger having $N$ inputs is a logic cone consisting of \texttt{AND} gates and \texttt{NOT} gates with one output net that activates the Trojan when a particular $N$-bit combination appears at its inputs. A sequential trigger is a sequence detector consisting of logic gates and flip-flops that activates the Trojan when a specific sequence appears on the input net of the trigger at consecutive clock cycles. Example triggers can be seen in Fig.~\ref{fig:trigger}. These trigger types form the basis of nearly all published HT triggers. The trigger may be implanted anywhere in the circuit. \iffalse It can be near the inputs or outputs or buried deep in the logic of the circuit.\fi %The trigger may be placed in any part of the chip, including, but not limited to the processor, memory, power management unit and input/output modules.

%% file: background.tex
%\nikhilc{Transferred from DATE paper, need to fix / adapt changed methodologies / remove irrelevant parts.}
We now outline the data generation methodology adopted for our HT trigger detection task. Each benign circuit (i.e., a circuit lacking a HT) is converted to a compact adjacency matrix representation and randomly embedded with triggers. The inverse node fanins for each node of the resulting circuit (potentially embedded with triggers) is extracted and saved for use by a learning model for HT detection. Data generation is discussed below.

\begin{figure}[tb]
  \centering
\includegraphics[scale=0.8]{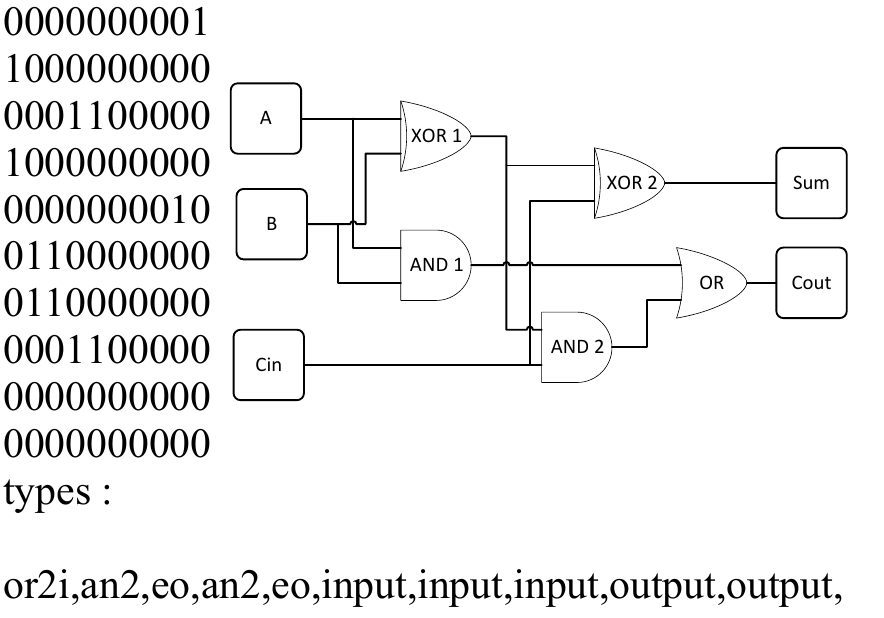}
  \caption{Full adder represented as a circuit adjacency matrix.  The circuit is shown as gates on the right here as an illustration, this is not typically included in a circuit adjacency matrix.}
  \label{fig:amFullAdder}
\end{figure}

\textbf{Circuit Adjacency Matrix}: For a synthesized circuit $C$, \iffalse synthesized to $V$ cells,\fi a Circuit Adjacency Matrix $A\in \mathbb{B}^{|V|\times |V|}$ is binary, with $V$ nodes, and represents $C$ as a graph. A node corresponds to a cell (logic gates and flipflops), input or output of a circuit. The elements of $A$ indicate which nodes in $C$ are connected to each other. \iffalse Along with an adjacency matrix, a list of nodes is required to preserve the functionality of the circuit.  This is included in the circuit adjacency matrix with the first node in the list corresponding to the first row and column in the matrix, the second node corresponding to the second row and column, and so forth.\fi In addition to $A$, we also maintain a list of standard cell type corresponding to each node such that the $i^{th}$ element in the list corresponds to the standard cell type of the cell represented in $i^{th}$ row and $i^{th}$ column of $A$. No data is lost in representation of a circuit adjacency matrix as the entire circuit can be reproduced from it.  An example adjacency matrix can be seen in Fig.~\ref{fig:amFullAdder}. It depicts a full adder represented as a circuit adjacency matrix, and the gate representation of the circuit is shown beside it for reference. The types or2i, an2 and eo refer to the standard cells \texttt{OR}, \texttt{AND} and \texttt{EXOR}.

\textbf{Trigger Generation}: Combinatorial and sequential triggers are generated as follows.  For a combinatorial trigger, we specify the number of inputs desired. Then we create a tree structure starting with a single gate as the root node and continuing until the desired number of inputs has been achieved. Each node is randomly selected as either an \texttt{AND} gate, a \texttt{NOT} gate or an \texttt{INPUT}. \texttt{AND} gates have two child nodes, while \texttt{NOT} gates have one, and \texttt{INPUTS} are leaf nodes. When each leaf node is an \texttt{INPUT} the trigger structure is complete. Trees with incorrect number of \texttt{INPUT} nodes or ones where not all leaf nodes are \texttt{INPUT}, are discarded.
\iffalse 
\begin{itemize}
    \item Specify the number of inputs desired for the trigger.
    \item Create a tree structure starting with a single gate as the root node and continuing until the desired number of inputs has been achieved.  Each node will be randomly selected as either an \texttt{AND} gate, a \texttt{NOT} gate or an \texttt{INPUT}.  \texttt{AND} gates have two child nodes, while \texttt{NOT} gates have one, and \texttt{INPUTS} are leaf nodes.
    \item When each leaf node is an \texttt{INPUT} the trigger structure has been finalized. Trees with incorrect number of \texttt{INPUT} nodes or ones where not all leaf nodes are \texttt{INPUT}, are discarded.
    \iffalse In the event that a tree structure does not achieve the correct number of \texttt{INPUT} nodes or not all leaf nodes are an \texttt{INPUT}, disgard the tree and start again.\fi 
\end{itemize}
\fi 
In order to generate sequential triggers, we compile a script, which for a given sequence, generates a Verilog HDL code for a non-overlapping Mealy sequence. This HDL code is then synthesized to a gate-level netlist and then converted to circuit adjacency matrix.

Triggers thus generated are structurally and functionally representative of different types of combinatorial and sequential triggers that can be used in trigger-based HT \cite{vosatka2018introduction}. For a combinatorial trigger, trigger size is equal to the number of inputs of the trigger and for a sequential trigger, it refers to the number of flipflops in the trigger circuit (equivalent to base-2 logarithm of input sequence length). 

\textbf{Trigger Embedding}: Circuit adjacency matrices are then embedded with triggers by appropriately adding rows and columns equivalent to number of trigger nodes and creating appropriate connections between trigger inputs and randomly selected insertion nodes of the benign circuit adjacency matrix. \iffalse Some were attached directly to inputs while others were dispersed throughout the circuit;\fi \iffalse All insertion points are randomly selected for each trigger.\fi In this way, each adjacency matrix is embedded with each trigger to yield multiple HT embedded adjacency matrices.

\textbf{Inverse Node Fanin}: These trigger-embedded circuit adjacency matrices are used as the starting point to extract inverse node fanins (INF). The INF of a node $i$ in $A$ is a subgraph consisting of all paths starting at a circuit input and terminating at $i$.
\iffalse  We say that for each node in a circuit an inverse node fanin exists and is represented beginning with the node (say node $i$) itself followed by the nodes it is connected to with incoming edges. This process is recursively repeated until the inputs of the circuit are reached and the resulting subgraph is termed the \emph{inverse node fanin} of node $i$.\fi 
The INF for a node is an effective representation of its neighborhood. An entire circuit can be described by specifying the INF for each node in the circuit. \iffalse This is a feature rich representation which contains as much data as the adjacency matrix but can be viewed one node at a time, making it a good candidate for a form of data to supply to the deep learning model. In order to prepare datasets for deep learning, the inverse node fanin data was converted to a one-hot encoding scheme to represent the type of each node.  In order to preserve structure, all zeros in the one-hot scheme represents the lack of a gate.  Each gate with only a single input, such as a \texttt{NOT}, will have a node in front of it, as well as a one-hot encoding representing the lack of a second node.\fi INFs for all circuit nodes are generated in a straight-forward manner from circuit adjacency matrices. For a graph of an HT-embedded circuit, we define \textit{trigger INF} as the INF of the node corresponding to the output of the trigger and \textit{benign INFs} as the INFs of all the other nodes in the graph. Note that some benign INFs may contain a few trigger nodes but never the complete trigger.  \iffalse The adjacency matrices derived directly from the HDL provide inverse node fanin data that represent nodes that do not contain a hardware trojan trigger.\fi 

%\nikhilc{Abdullah: Add a line describing what fanins are considered triggers and what fanins are benign. Also mention that benign INF might consist of partial triggers}.

%% file: problem_formulation.tex
\input{gcn}

%% file: gcn.tex
\begin{figure*}[t]
    \centering
    \includegraphics[scale=0.25]{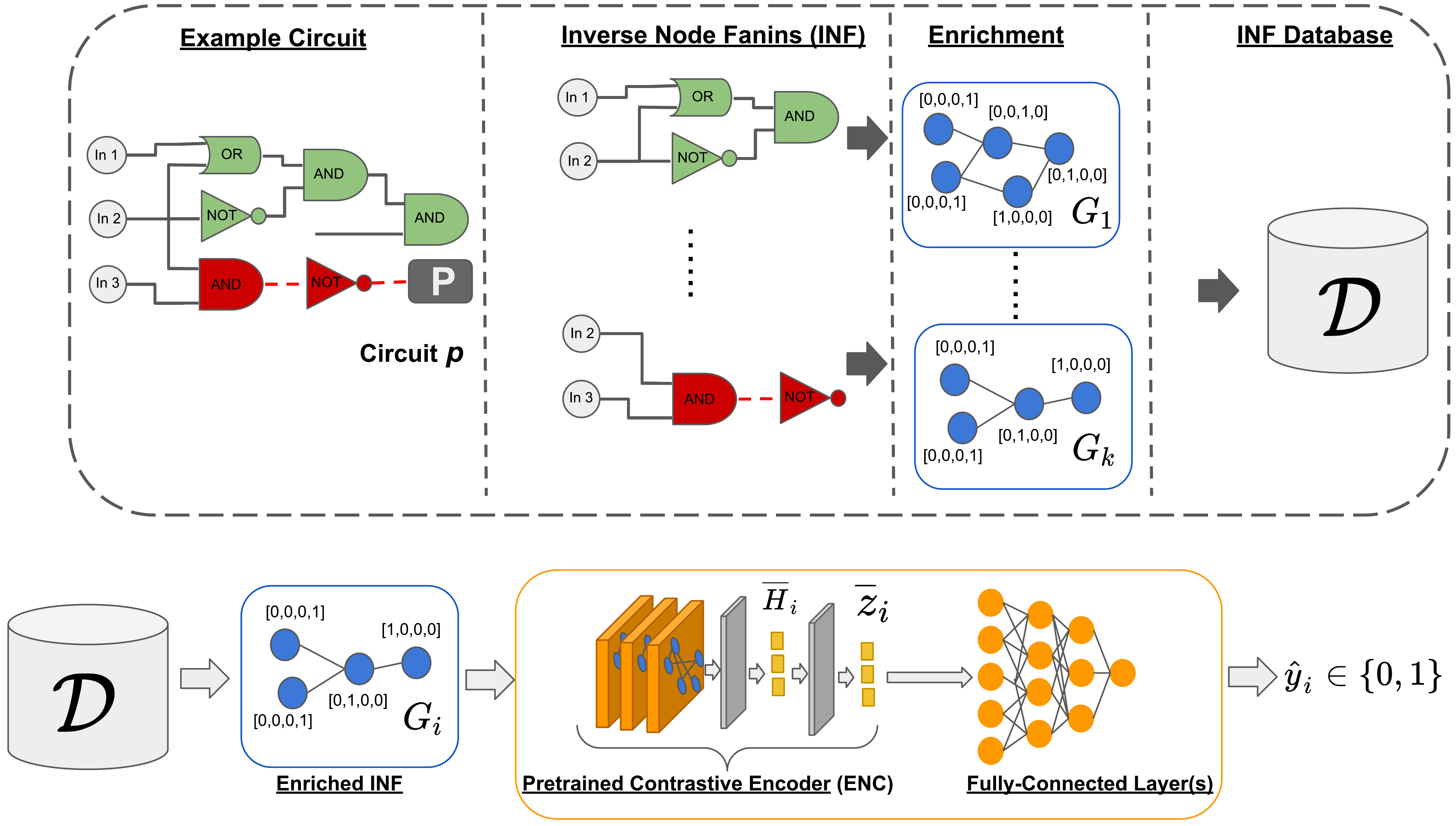}
    \caption{~\ourmethod: \ourmethodfull \iffalse HT Trigger Detection with Graph Convolutional Networks\fi . Here, we illustrate the classification pipeline with the help of an example circuit. A circuit $p$ with $k$ cells in it's netlist is broken down into $k$ INFs. Each INF is enriched with cell-types (one-hot encoding). Enriched INFs are represented as INF database $\mathcal{D}$. Each INF in $\mathcal{D}$ is passed into the classification model (orange box), wherein a latent representation $\bar{z}_i$ is obtained from the pretrained contrastive encoder (\emph{ENC}). $\bar{z}_i$ is then supplied to a fully connected network (FCN) to obtain the final classification $\hat{\mathrm{y}}_i\,$ for $G_i$. The classification model consists of a contrastive encoder (\emph{ENC}) (detailed in section.~\ref{sec:contrastive_encoder}) and a set of fully connected layers. The proposed classification pipeline is hence dependent ONLY on information available in the netlist and no further feature engineering is required.}
    \label{fig:gcn_pipeline}
\end{figure*}

A synthesized circuit can be viewed as a directed acyclic graph, hence we exploit GCN models which have demonstrated effectiveness in the context of unstructured data for  effective HT trigger detection. Details about the GCN learning mechanism are provided in~\cite{kipf2016semi}.

Let us consider a dataset $\mathcal{C} = \{C_1,C_2,...,C_N\}$ of hardware circuits, with each circuit potentially having a HT with a trigger and a payload embedded. Let us consider a circuit $C_p \in \mathcal{C}$; an HT embedded in $C_p$ comprises the trigger $t_p$ connected at one or more nets in $C_p$. Trigger $t_p$ is activated by a specific input pattern. Upon activation, $t_p$ activates the attached payload. We wish to detect the presence of HT in $C_p$ by training a learning model to detect $t_p$. Our learning model is trained \emph{only} on the circuit's netlist and we do not use any other features other than the gate-types of each gate in $C_p$. This deliberate restriction of our approach (i.e., to only use information available from the circuit netlist) is to ensure avoidance of cost associated with rich feature engineering pipelines that have been employed by previously proposed HT detection methods~\cite{hasegawa2016hardware,chen2017hardware,hoque18,liu2019hardware, kurihara}, preventing these methods from scaling to large circuit databases.  

Fig.~\ref{fig:gcn_pipeline} depicts the full classification pipeline for HT trigger detection that we employ. 
We first extract INF for all nodes from the adjacency matrix representation of $C_p$ as outlined in Section~\ref{sec:background}. Let $G^p_i$ be the INF of node $i$ for circuit $p$. $G^p_i$ is now considered a graph where each node (a cell in the original $C_p$) is enriched (tagged) with a one-hot vector indicating its cell-type. INF extraction and node-enrichment for each circuit in $C$ results in a dataset {$\mathcal{D} = \{G^1_1,..,G^1_{|V_1|},..,G^p_1,..,G^p_{|V_p|},..,G^N_{|V_N|}\}$}. Here, $|V_i|$ is the number of cells in $C_i$ (or nodes in $A_i$). Without loss of generality we can consider $\mathcal{D}$ to be a dataset of $m$ INFs $\mathcal{D} = \{G_1,...,G_m\}$ extracted from all circuits in $\mathcal{C}$. In this work, we consider each $G_i$ to be an undirected graph. Our~\ourmethod model is trained in a supervised manner to classify each graph $G_i$ as a benign or a trigger INF.

We can define the binary cross-entropy loss function to train~\ourmethod for the task of HT detection as
%
\iffalse 
\begin{equation}
    \small 
    \mathcal{L} = -\sum_{i=1}^m \mathrm{y}_i\mathrm{\textbf{ln}}(\mathrm{f}_\theta(\textbf{x}_i,\textbf{A}_i) + (1-\mathrm{y}_i)\mathrm{\textbf{ln}}(1 - \mathrm{f}_\theta(\textbf{x}_i,\textbf{A}_i))
    \label{eq:bce_loss}
\end{equation}
\fi 
\begin{equation}\label{eq:bce_loss} 
\small 
\begin{split}
    \mathcal{L} &= -\frac{1}{m}\sum_{i=1}^m \mathrm{y}_i\mathrm{\textbf{ln}}(\hat{\mathrm{y}}_i) + (1-\mathrm{y}_i)\mathrm{\textbf{ln}}(1 - \hat{\mathrm{y}}_i)\\
    \hat{\mathrm{y}}_i &= f_\Theta(\mathbf{x}_i,\mathbf{A}_i)
\end{split}
\end{equation}
where $f_\Theta$ represents the~\ourmethod learning model comprised of a set of parameters $\Theta = \{\theta^{(1)},\theta^{(2)},..,\theta^{(L)}\}$ where each $\theta^{(r)}$ represents parameters for layer $r$. $f_\Theta$ is a function of $\mathbf{A}_i \in \mathbb{B}^{l\times l}$ the adjacency matrix of $G_i$ (assume $G_i$ has $l$ nodes) and $\mathbf{x}_i \in R^{l\times 1}$, a vector of node features (indicating cell-type) for each node in $G_i$. Note, although we employ only one-hot encoding of cell-types in this work, $\mathbf{x}_i$ can quite easily be augmented to include richer features if available, thereby making~\ourmethod, a generic framework for representation learning for hardware circuits useful for various detection tasks.
\begin{figure*}[t]
    \centering
    \includegraphics[scale=0.28]{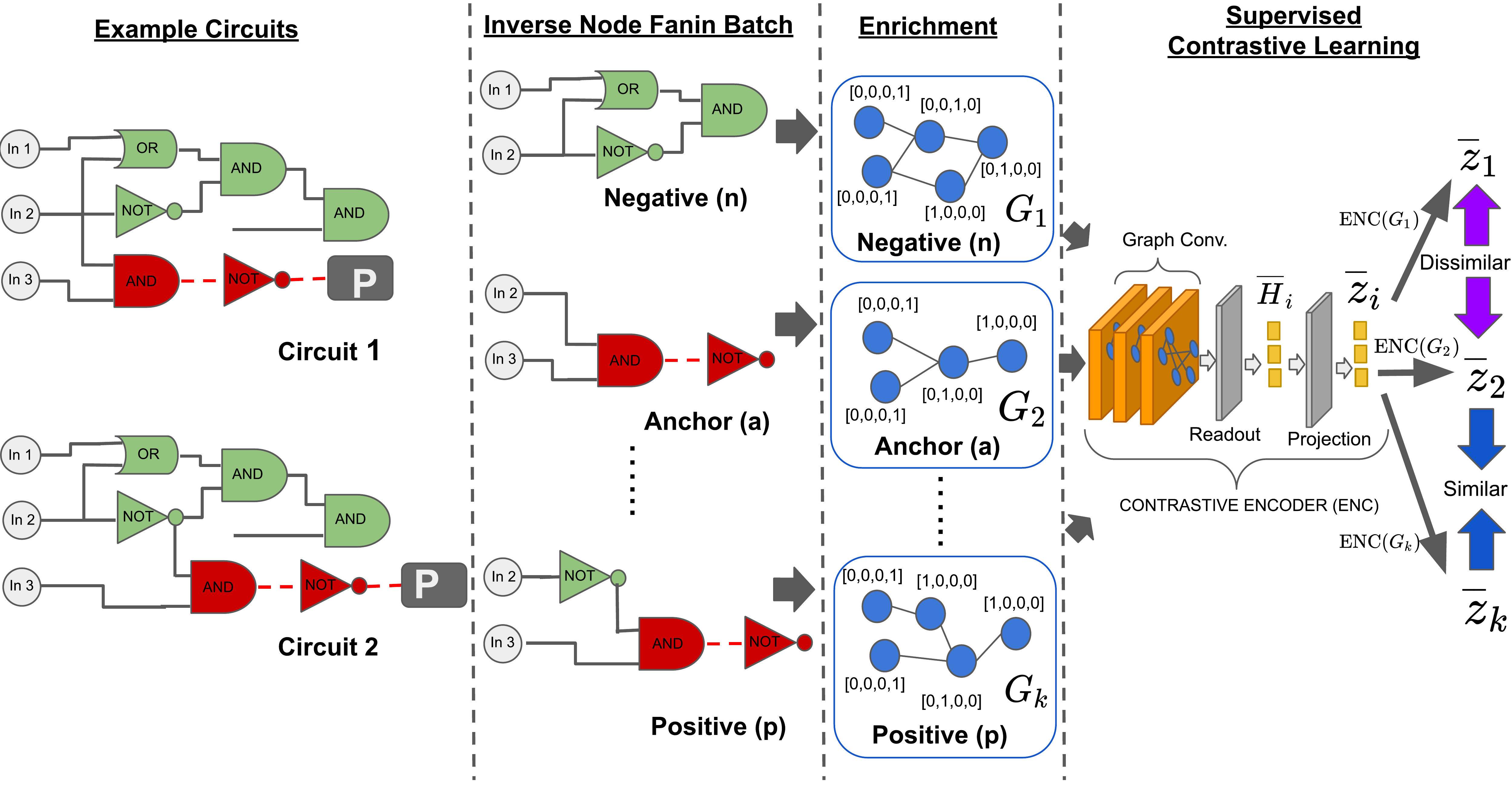}
    \caption{We illustrate the contrastive learning pipeline with an example scenario wherein $\mathcal{C}$ consists of only two circuits, each circuit broken down into INFs and corresponding enriched graphs to yield a total of $k$ INFs $\mathcal{D} = \{G_1,..,G_k\}$. Specifically, we consider the effect of contrastive learning for the fanin $G_2$ i.e., the trigger fanin from circuit 1 (which is called the anchor (a) in contrastive learning parlance). Here, fanin $G_k$ (positive (p) in contrastive learning parlance) has the same label (i.e., belongs to the same class) as $G_2$, hence the contrastive loss (Eq.~\ref{eq:contrastive_loss}) constrains $\overline{z}_2$ and $\overline{z}_k$ to be similar while causing $\overline{z}_1$ (representation for $G_1$; called negative (n) instance in contrastive learning parlance) to be dissimilar to both $\overline{z}_2$ and $\overline{z}_k$.}
    \label{fig:contrastive_learning}
\end{figure*}

\subsection{Supervised Contrastive Pretraining}\label{sec:contrastive_encoder}

Our~\ourmethod model consists of two parts (i) a \textit{contrastive encoder} (ENC) (inspired from the supervised contrastive learning model proposed by Khosla et al. in~\cite{khosla2020supervised} for image classification) followed by (ii) multiple fully connected layers (FCN). 

\textbf{Embedding:} The first part of ENC, consists of learning a latent embedding for the input graph $G_i$ which is achieved through three \textit{Graph Convolutional Network} (GCN) layers interleaved with ReLU activation functions~\cite{agarap2018deep}. Each GCN layer involves a topology-aware representation learning mechanism.
\begin{equation}
    \small 
    H^{(r+1)} = \mathrm{ReLU}(D_i^{\frac{1}{2}}A_iD_i^{\frac{1}{2}}H^{(r)}\theta^{(r)})
    \label{eq:gcn_layer}
\end{equation}
In Eq.~\ref{eq:gcn_layer}, $\mathrm{H}^{(r)}$ corresponds to the latent representation in the $r^{th}$ layer. $H^{(0)} = \mathbf{X}_i \in \mathbf{R^{l\times l}}$ (i.e., one-hot encodings for all nodes in $G_i$). $D_i$ is the degree matrix of $G_i$ and $\theta^{(r)}$ is the set of weights corresponding to layer $r$. The term $D_i^{\frac{1}{2}}A_iD_i^{\frac{1}{2}}$ indicates the symmetric normalized Laplacian matrix of $A_i$ and encodes the rich connectedness structure of the graph which is useful for propagating the node representations appropriately from one GCN layer to the next while also being employed in the message passing of node representations (details in ~\cite{kipf2016semi}).
Intuitively, the $r^{th}$ GCN layer learns node embeddings (for all nodes in $G_i$) influenced by neighbors up to $r$-hops away. 

Our~\ourmethod model consists of three GCN layers and hence considers effects of nodes up to 3-hops away. These GCN layers yield a set of 3 latent representations $\{H^{(1)},..,H^{(3)}\}$. The latent representation $H_3 \in \mathbb{R}^{l\times q}$ where $q$ represents the latent dimension and $l$ the number of nodes in $G_i$. $H_3$ is hence a matrix consisting of a q-dimensional latent representation for each node in $G_i$. A \textit{readout} step is employed to transform these individual node-representations present in $H_3$ into a single representation $\overline{H}_i \in \mathbb{R}^{q\times 1}$ for the entire graph $G_i$. In~\ourmethod, we employ mean pooling as the choice of \textit{readout} layer and hence our final graph representation $\overline{H}_i$ is an element-wise mean of all the individual node representations present in $H_3$.  

\textbf{Projection:} The second part of \textit{ENC} comprises a linear transformation of $\overline{H}_i$ (normalized to lie on the unit hypersphere) to obtain a \textit{projected} representation $\overline{z}_i \in \mathbb{R}^{q'\times 1}$, in our case $q'$ is the latent projection dimension set to 256. 
Hence, for each INF graph $G_i$ with $l$ nodes, the \textit{ENC} yields a latent representation $\overline{z}_i \in \mathbb{R}^{q' \times 1}$ projected onto the unit hypersphere.

\textbf{Contrastive Loss Formulation:} Most contrastive learning approaches proposed thus far~\cite{chen2020simple,you2020graph}, have been unsupervised (or more specifically ``self-supervised'') and due to the absence of explicit instance labels, rely on manipulating the input instance i.e., (cropping in case of images; adding or deleting nodes/edges in the case of graphs), and ensuring similar representations obtained for the original and manipulated instance. Overall, it has been found that contrastive learning approaches influence the learned representation (and improve model prediction quality) by lending greater structure to the latent space learned by the model.

In Fig.~\ref{fig:contrastive_learning}, an illustrative example of the effect of contrastive learning is presented. Let us consider a scenario with $\mathcal{C}$ comprising only of two circuits. Each circuit is broken down into its INFs and each node in every INF is enriched with its cell-type, as described previously in section~\ref{sec:problem_formulation} to obtain a dataset $\mathcal{D}$ of input graphs. Let us consider $\mathcal{D} = \{G_1,...,G_k\}$ to comprise of $k$ INFs from circuits in $\mathcal{C}$. Since each of the two circuits in $\mathcal{C}$, has a trigger embedded, each circuit has a single INF which is labeled a \textit{trigger} INF while the rest are benign INFs. 

Each $G_i \in \mathcal{D}$ is operated on by the contrastive encoder (ENC) in~\ourmethod, to obtain a set ${Z = \{\overline{z}_1,\overline{z}_2,..,\overline{z}_k | \overline{z}_i \in \mathbb{R}^{q'\times 1}\}}$ of projected representation vectors on the unit hypersphere. Owing to our supervised problem context, there also exist a set of corresponding labels $\mathrm{Y} = \{\mathrm{y}_1,..,\mathrm{y}_\mathrm{k}\}$ such that $\mathrm{y}_i \in \{0,1\}$.
Given, $\mathcal{D},\mathrm{Z},\mathrm{Y}$, our supervised contrastive loss is described in Eq.~\ref{eq:contrastive_loss}

\begin{equation}
    \small 
    \sum\limits_{G_i\in \mathcal{D}} \frac{-1}{|P(i)|} \sum\limits_{p \in P(i)} \mathrm{log} \frac{\mathrm{exp}(\mathrm{z}_{i} \boldsymbol{\cdot} z_p/\tau )}{\sum\limits_{G_j \in \mathcal{D}'(i)} \mathrm{exp}(z_i\boldsymbol{\cdot} z_j/\tau )}
    \label{eq:contrastive_loss}
\end{equation}
Here, for each graph $G_i \in \mathcal{D}$ (note $G_i$ is referred to as the \emph{anchor} graph), the set $P(i)$ includes all instances in $\mathcal{D}$ that have the same class label as $G_i$. Each member of $P(i)$ is considered a \emph{positive} instance (in contrastive learning parlance) w.r.t the \emph{anchor} instance $G_i$. The numerator of the logarithm in Eq.~\ref{eq:contrastive_loss}, calculates the dot product between the ENC generated latent representation of $G_i$ and ENC generated latent representation obtained for positive graph $G_p \in P(i)$. The denominator is a normalizing constant and is a summation of similar dot-products between $G_i$ and all other instances in $\mathcal{D}$ excluding $G_i$ (i.e., with all members in set ${\mathcal{D}'(i) = \mathcal{D}\setminus\{G_i\}}$). Intuitively, Eq.~\ref{eq:contrastive_loss} encourages representations $z_i$ belonging to the same class to be closer together while actively separating representations of different classes to be far apart. Hence, the goal of supervised contrastive learning coincides well with the general goal of classification pipelines which is to learn latent representations easily separable by a hyperplane between the various classes. 

\textbf{Pretraining:} Hence, we employ the aforementioned supervised contrastive learning procedure as a \emph{pretraining} step in our classification pipeline. Essentially this means, that the \emph{ENC} model is first trained using the contrastive loss function in Eq.~\ref{eq:contrastive_loss} after which the trained layers in \emph{ENC} are frozen (i.e., the weights are unaffected by backpropagation).

\textbf{Classification:} This frozen \emph{ENC} module is then combined with a 3-layer fully-connected network (FCN) which finally transforms $\overline{z}_i$ (i.e., each output of \emph{ENC}) to predict the probability of $G_i$ containing a trigger.

%% file: experimental_setup.tex
We now detail the data used for training and evaluation of~\ourmethod and also outline the evaluation procedure.
\subsection{Dataset Description}
\label{subsec:dataset}
In order to train \ourmethod to recognize trigger-based HTs, we required a database of circuits to be embedded with the triggers we generated. For this purpose, we use the open-source IP cores from Opencores.org \cite{opencores}. The cores are compiled and synthesized to flattenned gate-level netlists using a standard CMOS cell library \cite{djigbenou2007development}. Of all the available standard cells in the library, we use only the basic set of cells, comprising of logic gates: \texttt{AND}, \texttt{OR} and \texttt{NOT} and flipflops: D-Flipflop, D-Flipflop with asynchronous reset and D-Flipflop with asynchronous set and asynchronous reset. The restriction to a limited set of cells is done to have structural similarity with the HT triggers discussed in Section \ref{sec:background}, thereby, not making the detection easier. We used IP cores written in Verilog HDL and Synopsys DC compiler \cite{synopsysdc} as synthesis tools. GATE-Net only requires synthesized flattened netlist  Generally, our approach can be applied for netlist synthesis from any standard cell library and is independent of synthesis tools and HDL (Verilog/VHDL).
\begin{table}[!t]
    \centering
    \caption{Evaluation dataset: Types of IP cores and total cell count}
    \begin{tabular}{c|p{45mm}|c}\toprule
         No. & IP Core description & Total Num. cells\\  \hline
         
         1 & Antilogarithm function & 912\\ \hline
         2 & Logarithm function & 441\\ \hline
         3 & Cellular automata pseudo random number generator  & 931\\ \hline
         4 & Simple serial peripheral interface (SPI) & 968\\ \hline
         5 & Fixed point arithmetic module & 2362\\ \hline
         6 & Wishbone to LPC bridge & 723\\ \hline
         7 & Simon block cipher & 1443\\ \hline
         8 & Wishbone controlled FM transmission & 1445\\ \hline
         9 & Wishbone to AXI & 1355\\ \hline
         10 & Digital phased lock loop & 503\\ \hline
         11 & I2C slave & 908\\ \hline
         12 & Random number generator & 587\\ \hline
         13 & SPI-3 interface & 709\\ \hline
         14 & Vedic mathermatics & 827\\ \hline
         15 & USB host controller & 10781\\ \hline
         16 & Hight block cipher & 4534\\ \hline
         17 & Context adaptive variable length coding & 5407\\ \hline
         18 & Signed division & 4691\\ 
         \bottomrule
         
    \end{tabular}
    \label{tab:dataset_ipcores}
    \vspace{-0.75cm}
\end{table}
The set of IP cores used in evaluation and their size in terms of number of cells is shown in Table \ref{tab:dataset_ipcores}. Each IP core is a benign circuit, synthesized and converted to the adjacency matrix representation and a HT trigger is inserted at a random insertion point. For each benign circuit, multiple instances of trigger-embedded circuits are generated with a distinct trigger and at distinct insertion points. We further divide the dataset for our HT detection experiments into: a) a dataset consisting of benign circuits embedded with combinatorial triggers and b) a dataset consisting of benign circuits embedded with sequential triggers. The procedure outlined in Section~\ref{sec:background} is used to embed triggers of sizes 15--35 for combinatorial and 3--5 for sequential triggers. The INFs for both datasets are then extracted. Table~\ref{tab:dataset_statistics} details the distribution of benign and trigger INFs in each dataset. We retain the imbalanced distribution of benign and trigger INFs to maintain a realistic, challenging classification setting.

\iffalse each trained using circuit fanin embeddings obtained using the existence vector based method introduced by~\cite{daiconvcirc2017}.\fi \iffalse and detailed in sec.~\ref{sec:background} \nikhilc{Need to add short description of this EV based method in Background}.\fi 

\iffalse \abdullahc{Given limited computing resources (Intel Core i7 Processor, 15GB Memory), we were able to evaluate circuits containing upto 10,000 cells. With enterprise-level resources, GATE-Net will scale to larger circuits. Moreover, based on our implementations, GATE-Net has better scalability than SOTA approaches. This is because Hoque-et-al. and Kurihara-et-al. extract multiple features (14 and 11 respectively) per net, while GATE-Net extracts one INF per net.}\fi  We plan to publicly release a larger dataset, comprising of 58 IP cores, and comprehensive in terms of a) circuit sizes: the number of  cells in synthesized gate-level netlists ranges from 200 to 200000; and b) application area: arithmetic, communication, and encryption, among others. 
\iffalse We use a representative subset of cores,  comprising 18 different circuits for our experiments, including, but not limited to, communication core such as USB controller, or an arithmetic core of phase-locked loop implementation; further details in Appendix~\cite{appendix}.\fi \iffalse processor, arithmetic, memory, power management, bus interfaces etc., and c) application areas: digital signal processing, cryptography, communication etc.\fi 
\iffalse \abdullahc{Need to include a bar graph for IP cores displaying application areas and sizes in the appendix.}\fi 
\iffalse Using the procedure outlined in Section~\ref{sec:background}, each circuit is converted to it's adjacency matrix form, embedded with triggers and then has the inverse node fanins extracted for each of it's node.\fi

\begin{table}[!t]
    \centering
    \caption{Dataset Statistics}
    \begin{tabular}{c|c|c|c}\toprule
         Dataset&Experiment Type&  Benign Fanins& Trigger Fanins\\  \hline
         
         \multirow{2}{*}{Comb. Triggers}&Train &2161 & 1485 \\ \cline{2-4}
         &Test & 2222 & 232\\ \hline\hline
        %  \multirow{2}{*}{Seq. Triggers}&Train &8246 &2270  \\ \cline{2-4}
        %  &Test &8562 & 4500\\ \bottomrule
        
        \multirow{2}{*}{Seq. Triggers}&Train & 1145& 723\\ \cline{2-4}
         &Test & 2977& 2470\\ \bottomrule
         
    \end{tabular}
    \label{tab:dataset_statistics}
    \vspace{-0.5cm}
\end{table}
\iffalse
\par\noindent
\textbf{Trigger Embedding}:\nikhilc{Abdullah and Nikhil: Characterize trigger embedding process, add fanin length figures. Abdullah can add text for the actual embedding process. Nikhil can add figures corresponding to density of fanin lengths for trigger,benign fanins.}\fi 
\subsection{Testing Environments}
We use the following two environments to test and compare \ourmethod with existing state of the art (SOTA) approaches:

\textbf{Random-Shuffle Testing}: In this type of testing, we randomly assign INFs to the training and testing sets. This would imply the possibility of benign INFs from the same circuit present in both training and testing sets. However, the actual INFs are NOT identical as each node has a unique INF and the INF for each node in a circuit is used ONLY once either in training or testing (never both). 
\iffalse This would mean that the benign INFs derived from the same circuit can appear in both the sets but are not identical as an INF is unique for each node.\fi By the same reasoning, the trigger INFs will also be distinct in both the sets, where at best, the same trigger embedded in a different benign circuit can be encountered in both the sets.

\textbf{Extrapolation-Based Testing}: We also create a significantly more challenging testing scenario to gauge the power of \emph{strict} extrapolation of the models considered. Here, we ensure that INFs in training and testing sets arise from disjoint sets of circuits. Furthermore, a trigger seen during training is never encountered during testing (not even embedded in a different circuit); in particular, triggers of different sizes are employed for training and test to ensure this. The circuits \#1--\#5 and \#12--\#18 of Table \ref{tab:dataset_ipcores} are used in the training and the rest are used for testing. The circuits in training are embedded with combinatorial triggers of sizes 20--35 and sequential triggers of sizes 4,5 and circuits in testing with combinatorial triggers of sizes 10,15 and sequential triggers of size 3. \iffalse Thus, we ensure a strict extrapolation experiment which truly tests the generalization ability of an HT trigger detection model. Finally we used multiple circuit types in training and testing to promote effective learning by use of heterogeneous fanin data.\fi  \iffalse For both datasets, we use predominantly \textit{arithmetic}, \textit{communication} and \textit{encryption} circuits as benign circuits.\fi 

\subsection{Evaluation}\label{sec:evaluation}
\par\noindent
To inspect representation learning capability of~\ourmethod, we compare it's performance on the task of HT  detection with standard SOTA classifiers presented by Kurihara et al. \cite{kurihara} and Hoque et al. \cite{hoque18}. For this purpose, we use the dataset comprising of IP cores numbered 1--7 from Table \ref{tab:dataset_ipcores}, each embedded with 21 combinatorial triggers and 30 sequential triggers, thereby generating a total of 357 instances of trigger-embedded circuits (INF statistics in Table~\ref{tab:dataset_statistics_baselines}). For each of these instances, we extract various structural and functional features presented in \cite{kurihara} and \cite{hoque18}, shown in Table \ref{tab:baseline_features}.  

\begin{table}[!t]
    \centering
    \caption{Dataset statistics for comparison against state-of-the-art approaches. Due to the expensive feature engineering required for Hoque et al.~\cite{hoque18} and Kurihara et al.~\cite{kurihara}, models, it was intractable to evaluate them on the full dataset (Table~\ref{tab:dataset_statistics}) and we hence employ a smaller subset for SOTA comparison.}
    \begin{tabular}{c|c|c|c}\toprule
         Dataset&Experiment Type&  Benign Fanins& Trigger Fanins\\  \hline
         
         \multirow{2}{*}{Comb. Triggers}&Train & 3138 & 126 \\ \cline{2-4}
         &Test & 768 & 29\\ \hline\hline
        %  \multirow{2}{*}{Seq. Triggers}&Train &8246 &2270  \\ \cline{2-4}
        %  &Test &8562 & 4500\\ \bottomrule
        
        \multirow{2}{*}{Seq. Triggers}&Train & 3211 & 163\\ \cline{2-4}
         &Test & 788 & 47\\ \bottomrule
         
    \end{tabular}
    \label{tab:dataset_statistics_baselines}
\end{table}

\textbf{Kurihara et al. \cite{kurihara}}: This approach uses 11 structural features to classify whether a net is a HT net or a normal net. We have extracted these features for each net in our gate-level netlists. The exceptions being feature \#9, as the output of the trigger in our representation does not connect to the circuit, and feature \#11 because we do not use multiplexers in our synthesized gate-level netlist.

\textbf{Hoque et al. \cite{hoque18}}: This approach uses 5 structural features and 9 functional features to classify whether a net is a HT net or a normal net. The structural features, \#1--\#4 are extracted from the gate-level netlist for each net and feature \#5 is not used for the reason discussed above. The functional features, \#6--\#9 are extracted by synthesizing the HT-embedded circuits in the Xilinx Vivado tool by using the command \texttt{report\_switching\_activity} for each net. As only 2-input logic gates and D-flipflops are used in our approach, feature \#10 is set to 1 for all nets. The controllability-based features, \#11--\#14 are not considered because the control value based identification, derived from boolean functional analysis, has limitations against HTs with sequential triggers as they are triggered by an input stream over a period of time \cite{waksman2013fanci}. 

The features, thus extracted, are assigned to each node in the adjacency matrix representation and the INFs are generated. Each node in the INFs is then passed on to the classifiers (used in \cite{kurihara} and \cite{hoque18} respectively). If more than 5\% of the nodes in an INF are classified as trigger nodes, then the INF is detected as trigger INF.

%We employ strong classification baselines detailed in~\cite{hastie2009elements}. (1) Logistic Regression (Log. Reg.) (2) Decision Tree (DT) (3) Random Forest (RF) with 100 trees. (4) XGBoost (XGB)~\cite{chen2016xgboost} with 10 trees (best XGB variant). (5) Support Vector Machine (SVM-RBF) with a radial-basis function kernel. (6) Multi-Layer Perceptrion (MLP).  
\iffalse 
\begin{enumerate}
    \item Logistic Regression (Log. Reg.): A standard classification model frequently applied in binary classification setting.s
    \item Decision Tree Classifier (DT): Another standard classification model, prone to high variance in classification performance.
    \item Random Forest (RF): An ensemble classification model based on DT, geared toward reducing variance of DTs via. the ensemble. We employed RF with 100 estimators which yielded best performance.
    \item XGBoost (XGB): A state-of-the-art classification model based on gradient-boosting principle, also an ensemble model based on DTs. We found that XGB with 10 estimators was the best performing in our experiments.
    \item Support Vector Machine (SVM-RBF): The SVM model is popularly used in many HT detection applications. We employ the radial-basis function (RBF) kernel in the SVM.
    \item Multi-Layer Perceptron (MLP): A fully-connected network classification model with three layers (64,128,256), trained to run for 50 epochs without early stopping.
\end{enumerate}
\fi 

\begin{table}[!t]
    \centering
    \caption{Structurual and Functional Features used in the State of the Art approaches}
    \begin{tabular}{p{8mm}|p{38mm}|p{28mm}}\toprule
         No. & Kurihara et al. \cite{kurihara} & Hoque et al. \cite{hoque18}\\  \hline
         
         1 &  No. of immediate fan-in upto 4-level away from input side of net & No. of immediate fan-in of net\\ \hline
         2 & No. of immediate fan-in upto 5-level away from input side of net & No. of immediate fan-out of net\\ \hline
         3 & No. of flipflops upto 4-level away from input side of net & Cell type driving the net\\ \hline
         4 & No. of flipflops upto 3-level away from output side of net & Min. distance from primary input\\ \hline
         5 & No. of flipflops upto 4-level away from output side of net & Min. distance from primary output\\ \hline
         6 & No. of 4-level loops on the input side of net & Static probability\\ \hline
         7 & No. of 5-level loops on the output side of net & Signal rate\\ \hline
         8 & Min. levels to primary input & Toggle rate\\ \hline
         9 & Min. distance to primary output & Min. toggle rate\\ \hline
         10 & Min. levels to nearest flipflop & Entropy of the driver function\\ \hline
         11 & Min. levels to nearest multiplexer & Lowest controllability of inputs\\ \hline
         12--14 & -- & Highest, average and standard deviation of controllability of inputs\\ 
         \bottomrule
         
    \end{tabular}
    \label{tab:baseline_features}
    \vspace{-0.4cm}
\end{table}

\iffalse 
\begin{table}[!t]
    \centering
    \caption{Structurual and Functional Features used in the State of the Art approaches}
    \begin{tabular}{p{39mm}|p{39mm}}\toprule
         Kurihara et al. \cite{kurihara} & Hoque et al. \cite{hoque18}\\  \hline
         
         No. of immediate fan-in upto 4-level away from input side of net & No. of immediate fan-in of net\\ \hline
         No. of immediate fan-in upto 5-level away from input side of net & No. of immediate fan-out of net\\ \hline
         No. of flipflops upto 4-level away from input side of net & Cell type driving the net\\ \hline
         No. of flipflops upto 3-level away from output side of net & Min. distance from primary input\\ \hline
         No. of flipflops upto 4-level away from output side of net & Min. distance from primary output\\ \hline
         No. of 4-level loops on the input side of net & Static probability\\ \hline
         No. of 5-level loops on the output side of net & Signal rate\\ \hline
         Min. levels to primary input & Toggle rate\\ \hline
         Min. distance to primary output & Min. toggle rate\\ \hline
         Min. levels to nearest flipflop & Entropy of the driver function\\ \hline
         Min. levels to nearest multiplexer & Lowest controllability of inputs\\ \hline
         -- & Highest, average and standard deviation of controllability of inputs\\ 
         \bottomrule
         
    \end{tabular}
    \label{tab:baseline_features}
\end{table}
\fi 

\textbf{Evaluation Metrics}: We employ standard classification metrics \emph{precision} (Prec), \emph{recall} (Rec), \emph{F1 score}~\cite{scikit-learn} for evaluating model performance. Intuitively, \emph{precision} penalizes the models for false positive classifications, i.e., classifying benign fanins as trigger fanins, while \emph{recall} represents the proportion of all trigger fanins correctly identified by the model. 
\begin{figure}[!tb]
    \centering
    \includegraphics[scale=0.23]{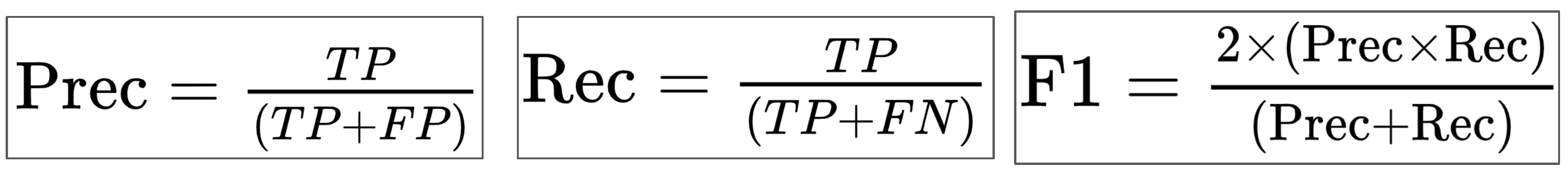}
    \caption{Evaluation Metrics.}
    \label{fig:eval_metrics}
    \vspace{-5pt}
\end{figure}
The \emph{F1 score} (range [0,1]) is the harmonic mean of precision and recall and a good indication of overall accuracy. In Fig.~\ref{fig:eval_metrics}, TP, FP, FN indicate true-positive (i.e., trigger INF that is correctly identified), false-positive (benign INF falsely identified as trigger INF) and false-negative (trigger INF falsely identified as benign INF) classifications respectively. \iffalse and yields a single number for holistic performance evaluation. A perfect model will yield a precision (prec.), recall (rec.) and F1 score of 1.0.\fi

%% file: results_and_discussion.tex
To evaluate the representation learning capability of~\ourmethod, we conduct a performance comparison on a downstream task of HT detection for a variety of hardware circuits embedded with combinatorial and sequential triggers. In each case, we compare~\ourmethod  with state-of-the-art classification models, namely Kurihara et al.~\cite{kurihara} and Hoque et al.~\cite{hoque18}. Each of the aforementioned state-of-the-art baseline HT detection approaches involve extensive feature-engineering and extract various structural and functional features of the circuit which are then employed in the HT detection task as described in section~\ref{sec:evaluation}. Our~\ourmethod model on the other hand only uses the graph structure of each INF (section~\ref{sec:problem_formulation}) enriched with cell-types (i.e., all information available directly form the netlist).

Specifically, we inspect three aspects to verify the effectiveness of our proposed~\ourmethod model:
\begin{enumerate}
    \item Performance of \ourmethod compared to state-of-the-art HT detection techniques for combinatorial and sequential triggers embedded in INFs.
    
    \item Effectiveness of contrastive learning of \ourmethod evaluated on a large expanded dataset comprising multiple trigger types and multiple groups of benign circuits, compared to a variant without contrastive learning for combinatorial and sequential INFs.
    
    \item Finally, we also characterize the performance of \ourmethod on an extremely challenging extrapolation setting for a holistic analysis of performance.
\end{enumerate}

\subsection{Comparison with State of the Art}
We characterize the performance of our proposed \ourmethod architecture relative to state of the art HT detection models proposed by Kurihara et al.~\cite{kurihara} in Table~\ref{tab:sota_kurihara_comparison}. The comparison is performed in the context of detecting combinatorial as well as sequential triggers embedded in hardware circuits. We notice that in both cases \ourmethod is able to outperform the model proposed by Kurihara et al.~\cite{kurihara} significantly. 
\begin{table}[!ht]
    \centering
    \caption{Comparison of \ourmethod with a state of the art HT detection model presented by Kurihara et al.~\cite{kurihara}.}
    \begin{tabular}{c|c|c|c|c}
        \toprule
        Expt. Type&Model & Precision & Recall & F1 \\ \hline
         \multirow{3}{*}{Combinatorial}&\ourmethod&\textbf{0.95}&\textbf{0.9}&\textbf{0.87}\\ \cline{2-5}
         &\ourmethod-noCont.&0&0&0\\\cline{2-5}
         &Kurihara et al.~\cite{kurihara}&0.83&0.4&0.54\\\midrule
          \multirow{3}{*}{Sequential}&\ourmethod&\textbf{1}&\textbf{0.92}&\textbf{0.96}\\ \cline{2-5}
          &\ourmethod-noCont.&\textbf{1}&0.5&0.67\\ \cline{2-5}
         &Kurihara et al.~\cite{kurihara}&0.68&0.87&0.76\\\bottomrule
    \end{tabular}
    \label{tab:sota_kurihara_comparison}
\end{table}
Specifically, we notice that our model outperforms theirs by \textbf{61.11\%} on the F1 score metric for combinatorial triggers and by \textbf{26.32\%} for sequential trigger detection, in both cases yielding significant performance improvement over the state of the art model. In the case of combinatorial trigger detection (which from the results may be inferred to be the harder of the two tasks), there is a significant deterioration in the recall of the Kurihara et al.~\cite{kurihara} model which is the main reason for performance degradation which indicates that this model is able to correctly identify only about 40\% of the triggers in the combinatorial fanin dataset. There is also a drop in the precision of the Kurihara et al.~\cite{kurihara} model for the sequential trigger detection task which is caused by the model detecting many false positives and from the table it may be inferred that the model is only correct 68\% of the time where it identifies an INF as containing a sequential trigger. Such a high false positive rate may be costly. \ourmethod on the otherhand is able to achieve very high precision (i.e., low false positive rates) and recall (i.e., able to identify all the variety of triggers) for both combinatorial and sequential trigger datasets.
\begin{table}[!tb]
    \centering
    \caption{Comparison of \ourmethod with a state of the art HT detection model presented by Hoque et al.~\cite{hoque18}}
    \begin{tabular}{c|c|c|c|c}
        \toprule
        Expt. Type&Model & Precision & Recall & F1 \\ \hline
         \multirow{3}{*}{Combinatorial}&\ourmethod&\textbf{0.96}&\textbf{0.9}&\textbf{0.93}\\ \cline{2-5}
         &\ourmethod-noCont.&0&0&0\\\cline{2-5}
         &Hoque et al.~\cite{hoque18}&0.94&0.55&0.7\\\midrule
          \multirow{3}{*}{Sequential}&\ourmethod&\textbf{1}&\textbf{0.89}&\textbf{0.94}\\ \cline{2-5}
          &\ourmethod-noCont.&0.95&0.79&0.86\\ \cline{2-5}
         &Hoque et al.~\cite{hoque18}&0.97&0.68&0.8\\\bottomrule
    \end{tabular}
    \label{tab:sota_hoque_comparison}
    \vspace{-0.4cm}
\end{table}

In Table~\ref{tab:sota_hoque_comparison}, we compare the \ourmethod model with another state-of-the-art HT detection model proposed by Hoque et al.~\cite{hoque18}, once again evaluating all models in the context of detecting combinatorial and sequential trigger INFs. We once again observe that \ourmethod is able to outperform the state-of-the-art model by Hoque et al.~\cite{hoque18} on both combinatorial and sequential HT deteciton tasks. However, unlike the Kurihara et al. model, we notice that the model proposed by Hoque et al.~\cite{hoque18} has high precision (i.e., low false positive rates) in both combinatorial and sequential HT detection cases. This may be attributed to the fact that the classification model in this case is actually an ensemble of three classifiers (i.e., Random Forest, Gradient Boosting, Naive Bayes) which is provably superior than the individual Random Forest model employed by Kurihara et al.~\cite{kurihara}. However, we notice that even in this case, the \ourmethod model outperforms the state of the art HT detection model by Hoque et al.~\cite{hoque18} by \textbf{32.86\%} for combinatorial HT detection and by \textbf{17.5\%} for sequential HT detection. Once again the deterioration in performance of Hoque et al.~\cite{hoque18} may be attributed to its inability to identify the variety of triggers resulting in the low recall value in both combinatorial and sequential HT detection contexts.

Finally, we also compared the \ourmethod-noCont. model which is a variant of the \ourmethod without the supervised contrastive pretraining phase. We observe some interesting trends consistent across Table~\ref{tab:sota_kurihara_comparison} and Table~\ref{tab:sota_hoque_comparison}. In both cases, we notice that the model is unable to recognize ANY combinatorial triggers. This may be attribured to the drastic data imbalance between the benign and trigger INFs during the training context of all models. It is interesting to note that contrastive learning in addition to allowing the model to learn better quality representations is also able to perform effectively in data imbalance contexts which is apparent from the results in both these tables by comparing performances of \ourmethod with \ourmethod-noCont. However, we also notice that \ourmethod-noCont. is able to yield a somewhat improved performance for the sequential HT detection task although still inferior to \ourmethod. We further investigate this comparative behavior between \ourmethod and its variant without contrastive learning in sec.~\ref{sec:contrastive_learning_effectiveness}. Note that although both Table~\ref{tab:sota_kurihara_comparison} and Table~\ref{tab:sota_hoque_comparison} use datasets derived from the one detailed in Table~\ref{tab:dataset_statistics_baselines}, due to the stochasticity in sampling of benign INFs for very large circuits (i.e., only INFs from a subset of benign nodes are sampled to make the INF extraction process tractable), the specific subsets of data may be slightly different although the overall data characteristics and experimental settings are identical across the two settings.

\subsection{Effectiveness of Contrastive Learning}\label{sec:contrastive_learning_effectiveness}
\begin{figure*}[!tb]
    \centering
    \begin{minipage}{0.2\textwidth}
        \includegraphics[scale=0.12]{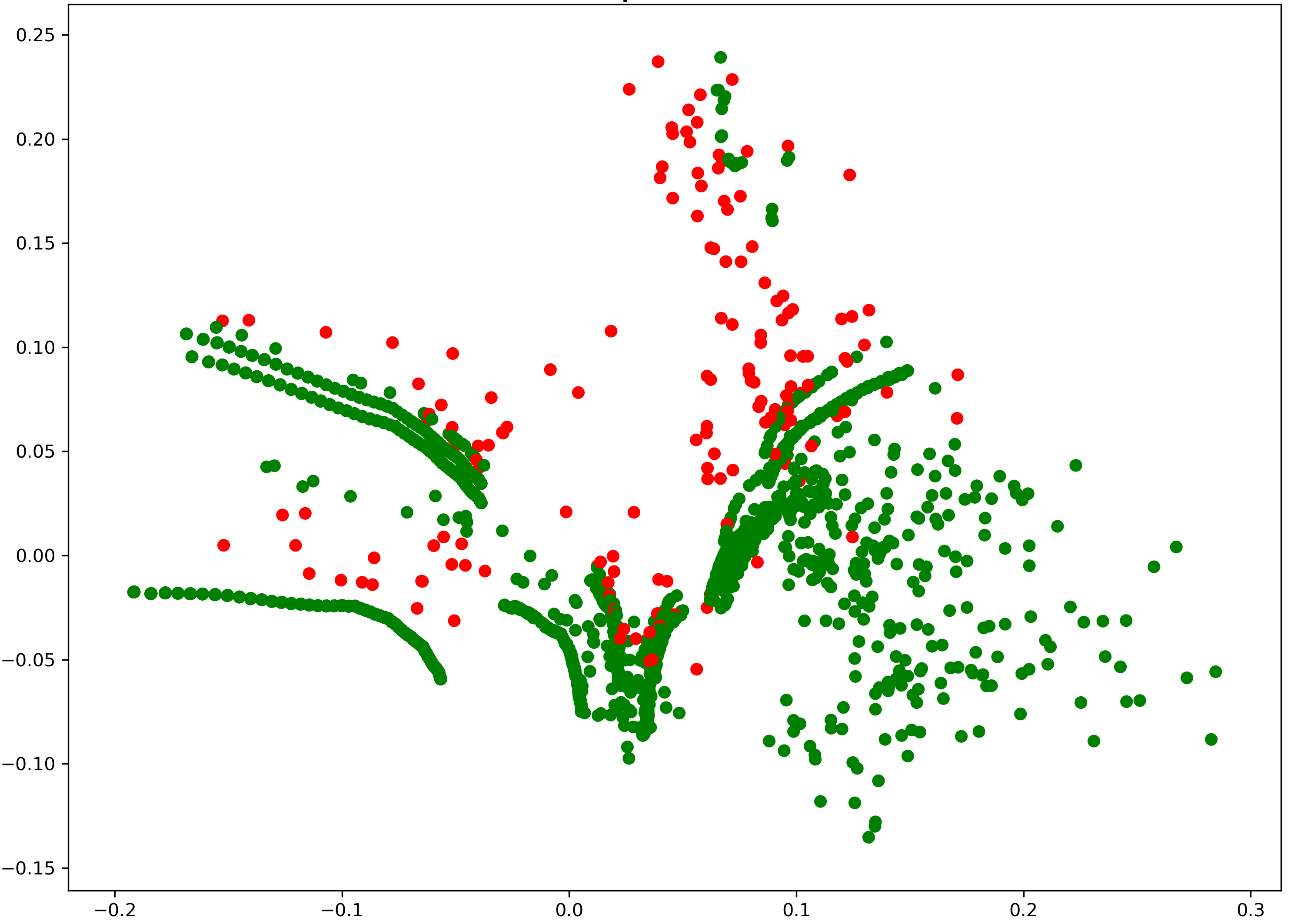}
        \subcaption{Epoch 0}
        \label{fig:pca_1}
    \end{minipage}
    \hfill
    \begin{minipage}{0.2\textwidth}
        \includegraphics[scale=0.12]{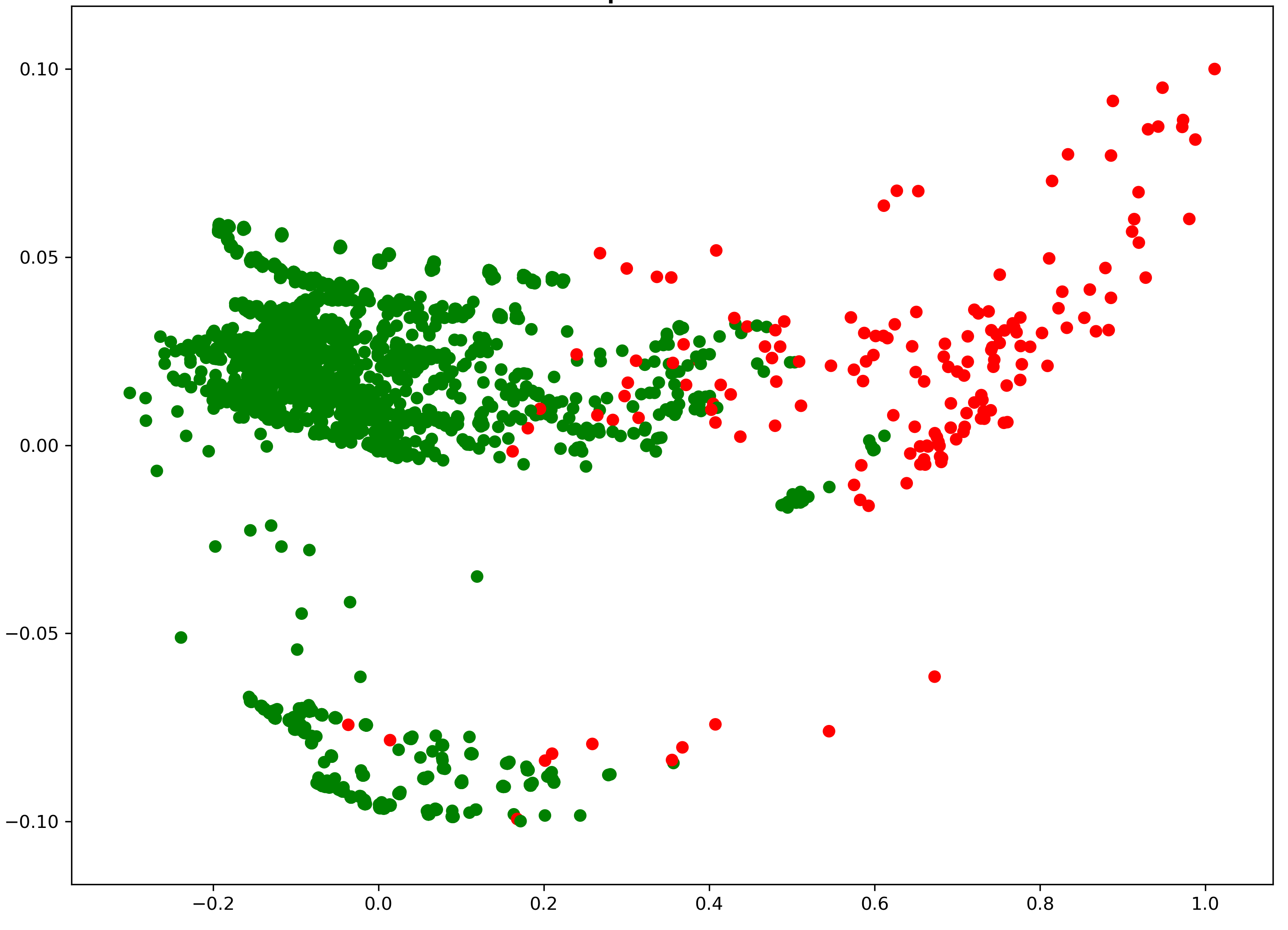}
        \subcaption{Epoch 50}
        \label{fig:pca_3}
    \end{minipage}
    \hfill
    \begin{minipage}{0.2\textwidth}
        \includegraphics[scale=0.12]{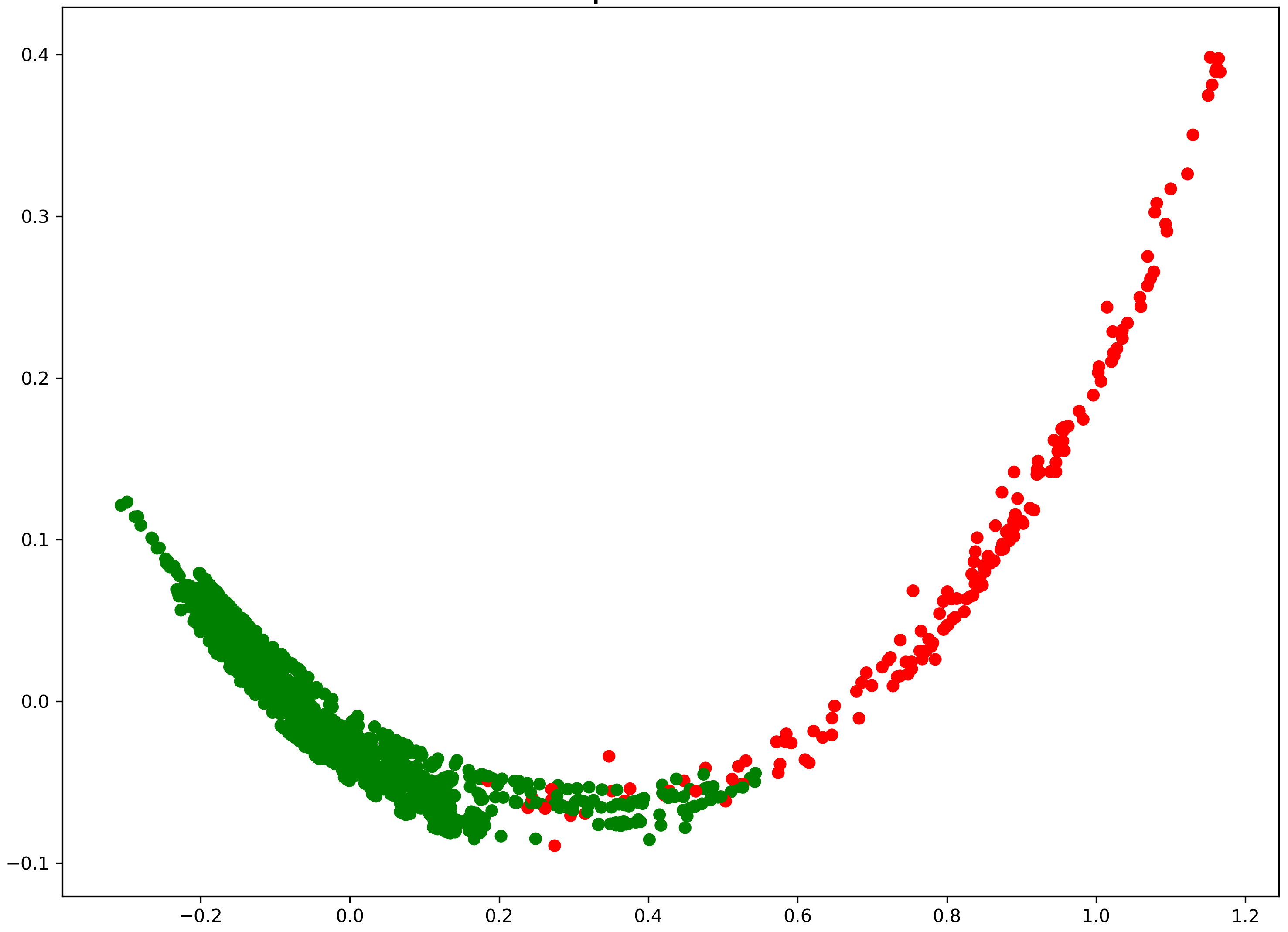}
        \subcaption{Epoch 100}
        \label{fig:pca_2}
    \end{minipage}
    \hfill
    \begin{minipage}{0.2\textwidth}
        \includegraphics[scale=0.12]{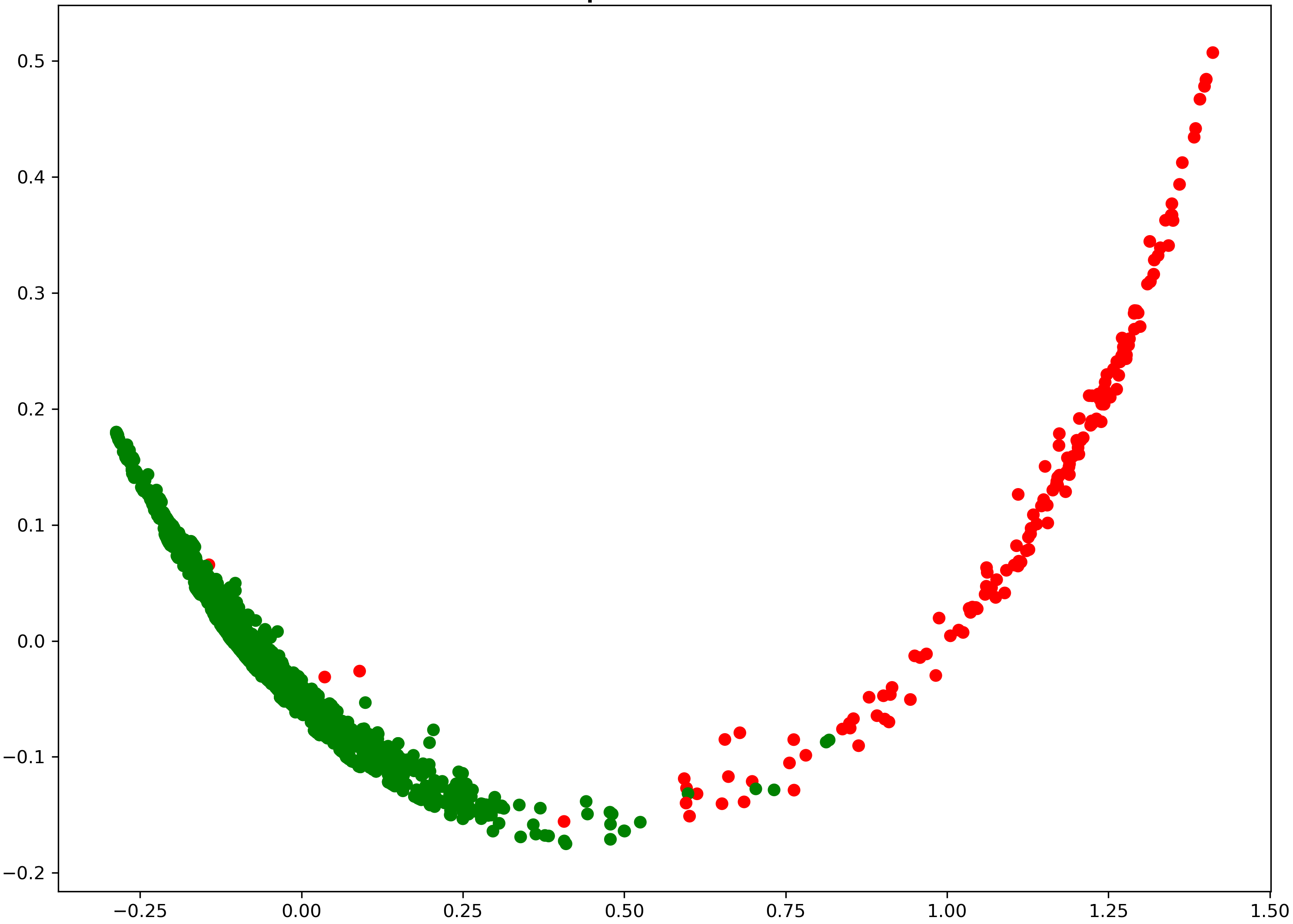}
        \subcaption{Epoch 150}
        \label{fig:pca_4}
    \end{minipage}
    \caption{Principal Component Analysis (PCA) reduced 2D embeddings of the representations $\overline{z}_i$ incrementally learnt by \ourmethod ENC model during contrastive pretraining over the course of 150 epochs. We notice that embeddings of benign INFs (green) and trigger INFs (red), start from a random projection setup before training at Epoch 0 (a), and are increasingly separated ultimately leading to the maximally separated representation seen in (d). This illustrates the effect of contrastive learning in the \ourmethod pipeline. Such explicitly separable embeddings are absent in \ourmethod-noCont. leading to inferior performance.}
    \label{fig:pca_contrastive_embeddings}
\end{figure*}

In order to test the effect of contrastive learning, we evaluate the \ourmethod and \ourmethod-noCont. models on much larger datasets for combinatorial and sequential HT detection. To simulate data paucity, we use only 40\% of the available data for training and the rest of the data is used for performance evaluation in this experiment. 
\begin{table}[!tb]
    \centering
    \caption{Effectiveness of Contrastive Learning in \ourmethod. We notice that \ourmethod outperforms variant without contrastive learning \ourmethod-noCont. in detection of both sequential and combinatorial triggers.}
    \begin{tabular}{c|c|c|c|c}
        \toprule
        Expt. Type&Model & Precision & Recall & F1 \\ \hline
         \multirow{2}{*}{Combinatorial}&\ourmethod&\textbf{0.99}&\textbf{0.98}&\textbf{0.99}\\ \cline{2-5}
         &\ourmethod-noCont.&0.93&0.72&0.81\\\midrule
          \multirow{2}{*}{Sequential}&\ourmethod&\textbf{0.96}&0.94&\textbf{0.95}\\ \cline{2-5}
         &\ourmethod-noCont.&0.9&\textbf{0.95}&0.92\\\bottomrule
    \end{tabular}
    \label{tab:contrastive_learning}
    \vspace{-0.4cm}
\end{table}
In Table~\ref{tab:contrastive_learning}, we notice that \ourmethod i.e., the model with constrastive learning (pre-training) outperforms the variant without constastive learning i.e., \ourmethod-noCont. We also notice that the performance of the variant of \ourmethod without constastive learning deteriorates significantly  for the combinatorial trigger detection task while the performance of \ourmethod with constrastive learning remains relatively consistent for both the trigger types, indicating that contrastive learning helps \ourmethod learn generalizable, robust representations of INFs for HT detection. We also notice that in the case of combinatorial trigger detection, the variant of \ourmethod without contrastive learning has low \emph{recall} which implies that it is unable to recognize many of the variety of triggers present in the dataset while \ourmethod (i.e., with constrastive learning) is able to \emph{recall} (i.e., correctly identify) 98\% of the triggers in the dataset further showcasing the power of contrastive learning to enable models to learn highly generalizable representations. We see that \ourmethod outperforms its variant without contrastive learning by \textbf{22.22\%} for combinatorial HT detection and by \textbf{3.26\%} for the sequential HT detection. Fig.~\ref{fig:pca_1}--\ref{fig:pca_4} qualitatively demonstrate the effect supervised contrastive pretraining has on the \ourmethod model representation thereby leading to superior performance over \ourmethod-noCont.
\begin{figure*}[!tb]
    \begin{subfigure}{.32\textwidth}
        \centering
        \vskip -0.07in
        \includegraphics[scale=0.22]{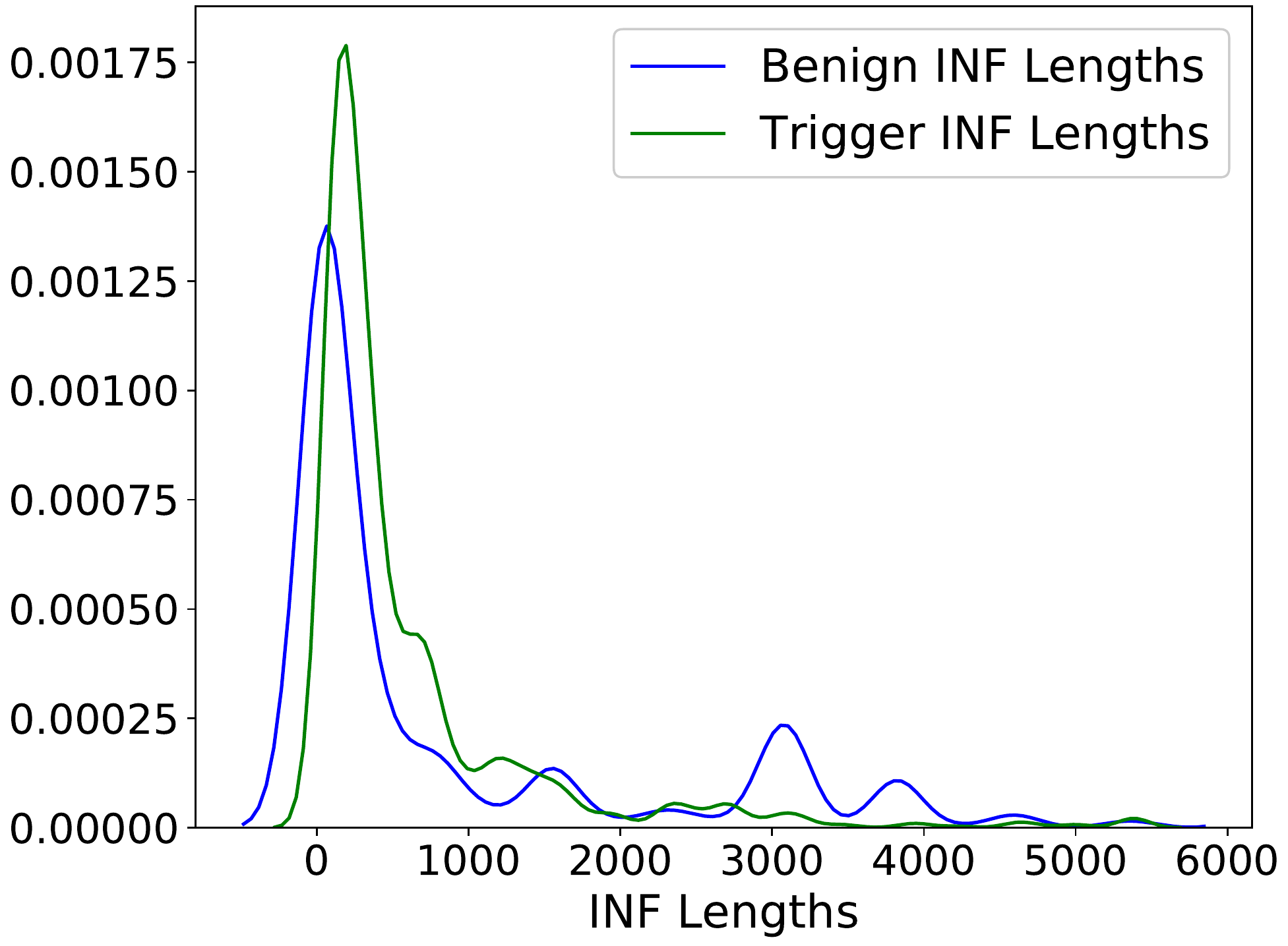}
        \vskip -0.05in
        \caption{}
        \label{fig:fanin_lengths_comb_random_embedding}
    \end{subfigure}
    \hfill
    \begin{subfigure}{.32\textwidth}
        \centering
        \includegraphics[scale=0.2]{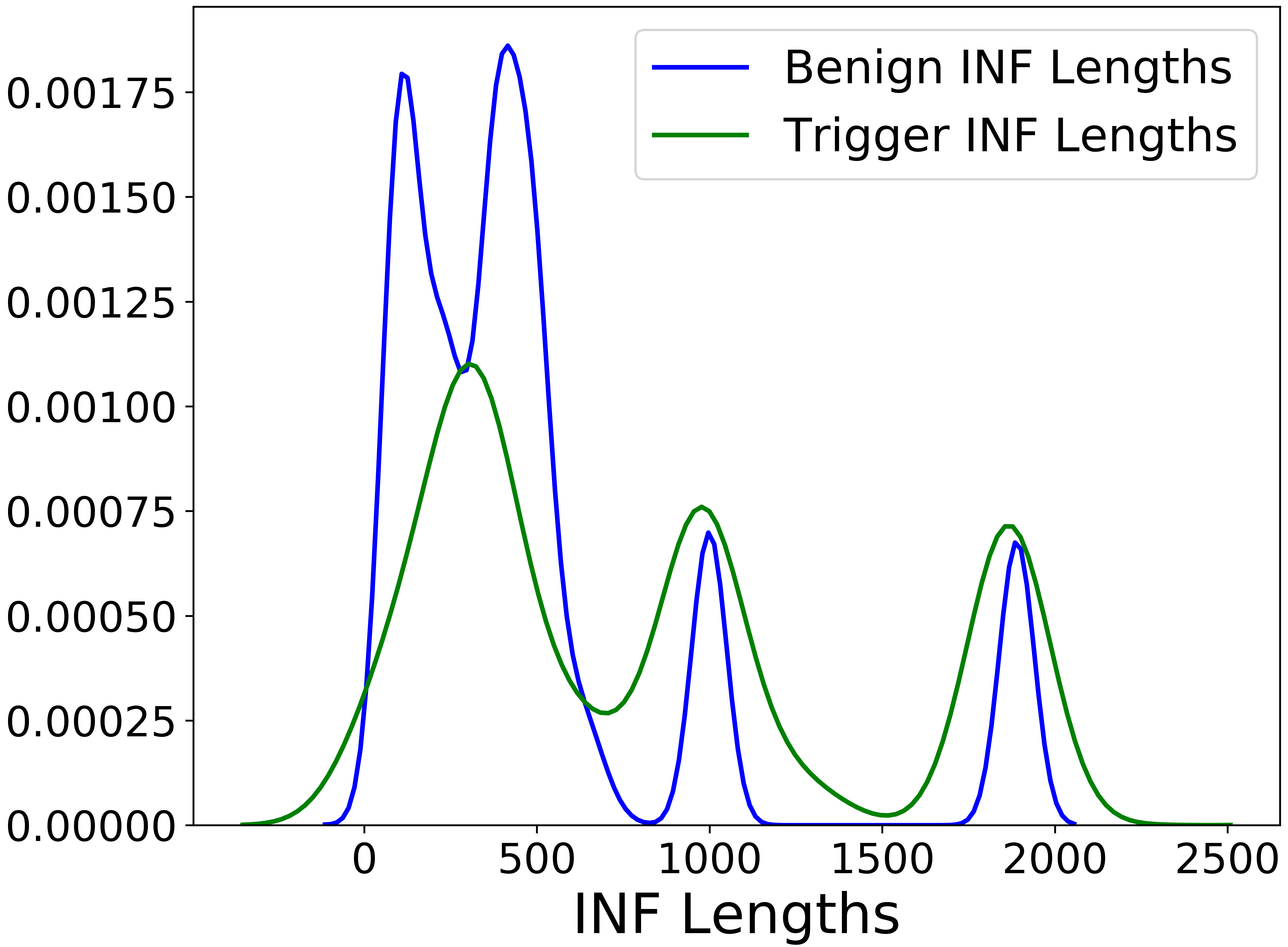}
        \caption{}
        \label{fig:fanin_lengths_comb_trig_shallow_embedding}
    \end{subfigure}
    \hfill
    \begin{subfigure}{.32\textwidth}
        \centering
        \includegraphics[width=1.5in,height=1.2in]{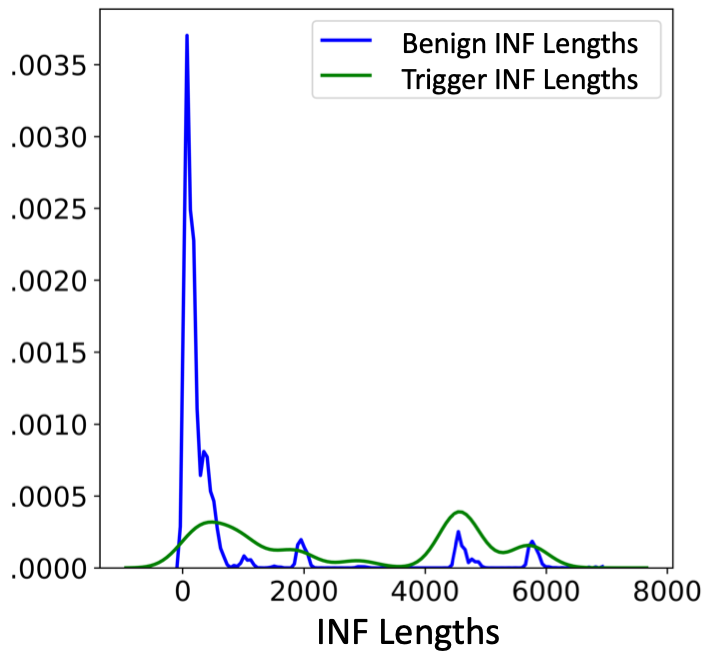}
        \caption{}
        \label{fig:fanin_lengths_seq_trig_shallow_embedding}
    \end{subfigure}
\label{fig:fanin_lengths}
\caption{(a) INF length distribution random trigger insertion on synthetic data consisting only of \texttt{AND,OR,NOT} gates and combinatorial triggers. (b) INF length distribution on real circuit data embedded with comb. triggers with shallow trigger embedding (c) INF length distribution on real circuit data embedded with seq. triggers with shallow trigger embedding.}
\vspace{-0.35cm}
\end{figure*}

\subsection{Extrapolation Performance}
We now evaluate the out-of-distribution generalization performance of \ourmethod, without explicitly training \ourmethod for this challenging experimental setting to understand out of the box behavior in this context and also to evaluate the effectiveness of contrastive learning in this context. To simulate out-of-distribution evaluation, we train our models only on a subset of the benign circuits and trigger sizes and evaluate model performance on triggers of different sizes embedded in completely different benign circuits.
\begin{table}[!tb]
    \centering
    \caption{Effectiveness of \ourmethod for out-of-distribution generalization compared with \ourmethod-noCont.}
    \begin{tabular}{c|c|c|c|c}
        \toprule
        Expt. Type&Model & Precision & Recall & F1 \\ \hline
         \multirow{2}{*}{Combinatorial}&\ourmethod&\textbf{0.38}&\textbf{0.67}&\textbf{0.48}\\ \cline{2-5}
         &\ourmethod-noCont.&0.3&0.66&0.42\\\midrule
          \multirow{2}{*}{Sequential}&\ourmethod&\textbf{0.54}&0.78&\textbf{0.64}\\ \cline{2-5}
         &\ourmethod-noCont.&0.51&\textbf{0.82}&0.63\\\bottomrule
    \end{tabular}
    \label{tab:out_of_distribution_generalization}
    \vspace{-0.3cm}
\end{table}
We notice from Table~\ref{tab:out_of_distribution_generalization} that both \ourmethod and \ourmethod-noCont. experience significant performance deterioration in the out-of-distribution extrapolation context. Specifically, the most significant deterioration occurs in the context of combinatorial HT detection which has been previously established to be the harder of the two detection tasks evaluated in this work. However, we still notice that in both cases, the F1 score of \ourmethod is higher than the variant without contrastive learning although the performance difference observed previously between the two models dwindled in the extrapolation setting inspected here. Our goal of evaluating \ourmethod in this challenging extrapolation context is not to display state-of-the-art performance results but rather to establish a baseline of performance of an out of the box model (\ourmethod) which has not been explicitly trained to excel in out-of-distribution generalization tasks so future efforts may build upon our pipeline to improve performance.

\par\noindent
\textbf{Results Summary}:~\ourmethod outperforms all SOTA classification models compared for HT detection. Hence, our results demonstrate superior representation learning capacity of \ourmethod, for HT detection. Overall, we achieve an average \textbf{46.99}\% performance improvement in F1 score over the two SOTA baselines for detection of combinatorial triggers and \textbf{21.91}\% improvement for detecting sequential triggers. We also inferred from the results that contrastive learning is very effective in allowing \ourmethod to learn robust, generalizable representations even in the context of data paucity. As apparent from Table.~\ref{tab:contrastive_learning}, \ourmethod achieves average performance improvement of \textbf{12.74\%} in F1 score across combinatorial and sequential HT detection.
\iffalse Finally, when the baseline classifiers are trained with INF embeddings generated by~\ourmethod, they outperform the corresponding variants trained using the existence vector based embeddings~\cite{daiconvcirc2017} by \textbf{21.05}\% on average on the HT detection task indicating that~\ourmethod yields richer circuit representations.\fi 
\par\noindent
\textbf{Security Analysis}: We now evaluate effects of various trigger embedding procedures and their effect on the degree of difficulty in HT detection. One approach is to randomly embed triggers, wherein a trigger with $k$ inputs has $k$ randomly chosen circuit nets connected to it. An experiment on a simple dataset of circuits (consisting only of \texttt{AND},\texttt{OR} and \texttt{NOT} gates) with randomly embedded triggers reveals interesting insights about the resulting INFs. In Fig.~\ref{fig:fanin_lengths_comb_random_embedding}, We observe a noticeable difference in the models of INF length distributions for both benign and trigger INFs. Also, INF lengths of size 2500 and above may be safely be regarded as benign (a trivial but effective detection strategy in this case). To avoid such biases in INF length, we assume a sophisticated attacker who has control over the trigger embedding process and adopts a \textit{shallow embedding} strategy wherein the trigger is embedded at most $k$-hops ($k = 2$ in our case) away from circuit inputs. This strategy alleviates discrepancy in INF length distributions between trigger and benign INFs, see Fig.~\ref{fig:fanin_lengths_comb_trig_shallow_embedding} and Fig.~\ref{fig:fanin_lengths_seq_trig_shallow_embedding} which both show much better agreement between INF length densities between trigger and benign INFs for combinatorial and sequential triggers respectively. \iffalse Notice that the shallow embedding strategy results in a multi-modal trigger INF length distribution with modes corresponding closely to benign INF length modes.\fi Triggers so embedded are harder to detect with trivial features like INF lengths.

%% file: conclusion.tex
In this paper we proposed~\ourmethod, a novel machine learning model based on supervised contrastive pre-training and graph convolutional networks for HT detection employing only data available from the circuit netlist. Through rigorous experimentation and comparison with several state-of-the-art baseline models we show that~\ourmethod achieves significantly better results for HT detection over a wide variety of circuit types with randomly embedded combinatorial (\textbf{46.99}\% performance improvement over baselines) and sequential triggers (\textbf{21.91}\% performance improvement over baselines). We also extensively evaluate the effect of supervised contrastive learning in \ourmethod and compare it with a variant of \ourmethod without contrastive learning and show qualitatively and quantitatively that supervised contrastive learning helps \ourmethod yield superior performance for HT detection.
We also proposed and detailed a methodology for generating and representing circuits embedded with HT triggers and have publicly released data and code for \ourmethod, trigger generation and embedding. We also performed analysis of out-of-distribution generalization performance of \ourmethod to serve as an effective baseline for future HT detection endeavors.  Finally, we performed a security analysis and detail advantages of a \emph{shallow} trigger embedding procedure for better trigger masking in the circuits. In the future, we wish to extend our current \ourmethod approach and characterize its performance in the context of adversarial trigger generation models.
\par\noindent
\textbf{Acknowledgements:} This work is supported in part by the National Science Foundation via grant DGE-1545362.